\RequirePackage{fix-cm}
\documentclass[twocolumn]{svjour3}          
\smartqed  
\usepackage{graphicx}
\usepackage{amsmath,amssymb}
\usepackage{mathptmx}
\usepackage{multirow}
\usepackage{marvosym}
\usepackage{booktabs}
\usepackage{subcaption}
\usepackage{bm}
\usepackage{rotating}
\usepackage{enumitem}
\usepackage{color}
\usepackage[pagebackref=true, colorlinks,bookmarks=false]{hyperref}

\usepackage[table,xcdraw]{xcolor}

\begin{document}

\title{Scene Prior Filtering for Depth Super-Resolution}


\author{Zhengxue Wang\textsuperscript{1}\textsuperscript{*} \and Zhiqiang Yan\textsuperscript{2}\textsuperscript{*} \and Ming-Hsuan Yang\textsuperscript{3} \and Jinshan Pan\textsuperscript{1} \and Guangwei Gao\textsuperscript{1}\textsuperscript{$\dagger$} \and Ying Tai\textsuperscript{4}  \and Jian Yang\textsuperscript{1,4 }\textsuperscript{$\dagger$}
\\
{\small {\{zxwang, jspan, gwgao, csjyang\}@njust.edu.cn}} ~~{\small yanzq@nus.edu.sg} ~~{\small mhyang@ucmerced.edu} ~~{\small yingtai@nju.edu.cn}}


\institute{
\textsuperscript{*} Equal contribution~~~
\textsuperscript{$\dagger$} Corresponding authors\\
\textsuperscript{1} PCA Lab, 
Nanjing University of Science and Technology, China.\\
\textsuperscript{2} National University of Singapore.\\
\textsuperscript{3} University of California, USA, and Yonsei University, South Korea.\\
\textsuperscript{4} PCA Lab, Nanjing University, China.}
\authorrunning{Zhengxue Wang, Zhiqiang Yan, Ming-Hsuan Yang, Jinshan Pan, Guangwei Gao, Ying Tai and Jian Yang}

\date{Received: date / Accepted: date}

\maketitle
 
\begin{abstract}
Multi-modal fusion serves as a cornerstone for successful depth map super-resolution.
However, commonly used fusion strategies, such as addition and concatenation, fall short of effectively bridging the modal gap. 
As a result, guided image filtering methods have been introduced to mitigate this issue. 
Nevertheless, it is observed that their filter kernels usually encounter significant texture interference and edge inaccuracy. 
To tackle these two challenges, we introduce a Scene Prior Filtering network, SPFNet, which utilizes the priors' surface normal and semantic map from large-scale models. 
Specifically, we propose an All-in-one Prior Propagation that computes similarity between multi-modal scene priors, \textit{i.e.}, RGB, normal, semantic, and depth, to reduce the texture interference. 
Besides, we design a One-to-one Prior Embedding that continuously embeds every single modal prior into depth using Mutual Guided Filtering, further alleviating texture interference while enhancing edge representations. 
Our SPFNet has been extensively evaluated on both real-world and synthetic datasets, achieving state-of-the-art performance. 
Project page: \url{https://yanzq95.github.io/projectpage/SPFNet/index.html}. 
\end{abstract}

\section{Introduction}
\begin{figure*}[t]
\centering
\includegraphics[width=0.9\textwidth]{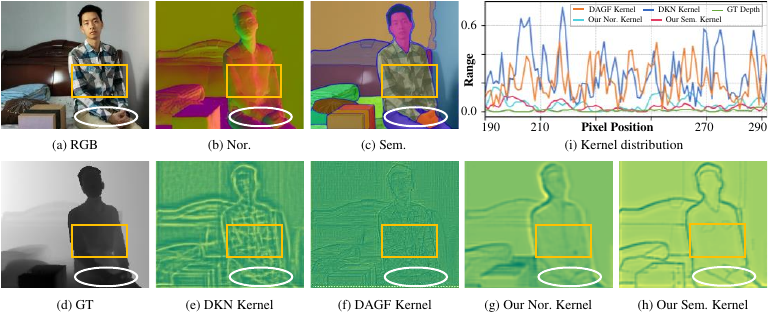}\\
\caption{\textbf{Visualizations of scene priors and filter kernels.} The normal (Nor.) (b) and semantic (Sem.) (c) are produced from (a) using Omnidata~\cite{eftekhar2021omnidata} and SAM~\cite{kirillov2023segment}, respectively. (e) and (f) are the filter kernels derived from DKN~\cite{kim2021deformable} and DAGF~\cite{zhong2023deep}, while (g) and (h) are from our SPFNet. (d) is the ground-truth (GT) depth, and (i) is the normalized kernel distribution.}\label{fig:FilterKernel}
\end{figure*}

Advances in sensor technology have led to the extensive application of depth cues in various fields, such as autonomous driving \cite{yan2023distortion,sun2021learning,qiao2024rgb}, 3D reconstruction \cite{song2020channel,yang2022codon,yan2023desnet}, and virtual reality \cite{yuan2023recurrent,zhou2023memory}. However, depth measurements are typically low resolution (LR) due to sensor limitations and the complexity of imaging environments. Recently, a number of guided image filtering approaches \cite{pan2019spatially,li2019joint,kim2021deformable,zhong2023deep} have been proposed to facilitate depth super-resolution (DSR). Nevertheless, the filter kernels, which are constructed directly from RGB images, often suffer from significant texture interference, and the clarity near edges is typically compromised. For instance, as depicted in Fig.~\ref{fig:FilterKernel}(e) and (f)  (yellow boxes), the filter kernels of DKN \cite{kim2021deformable} and DAGF \cite{zhong2023deep} contain a substantial amount of textures. Fig. \ref{fig:FilterKernel}(i) shows that these two kernels display too many abrupt changes, while the ground-truth depth is considerably smoother. These observations demonstrate that the texture interference is not conducive to depth recovery. Moreover, within the white boxes, the edges and their neighboring pixels show high similarity, leading to an insufficient contrast. 

Compared to the RGB input, Fig.~\ref{fig:FilterKernel}(b) indicates that the surface normal is largely devoid of texture interference. 
On the other hand, 
Fig.~\ref{fig:FilterKernel}(c) shows that the semantic map displays clear edges between different categories.
These inherent characteristics are highly advantageous for constructing filter kernels with less interference and more distinct edges. 

Drawing from the observations above, we design a novel scene prior filtering network (SPFNet) to reduce texture interference and enhance edge accuracy.
%
Specifically, we first employ large-scale models~\cite{eftekhar2021omnidata,kirillov2023segment} to generate normal and semantic priors from RGB input. 
An all-in-one prior propagation (APP) is introduced, which computes the similarity between multi-modal scene priors, \textit{i.e.}, RGB, normal, semantic, and depth, to weaken the interference. 
In addition, we present a one-to-one prior embedding (OPE) that sequentially incorporates each single-modal prior into depth using mutual guided filtering (MGF), further diminishing the interference and enhancing edges via normal and semantic. 
The MGF comprises a bidirectional path, that is, prior-to-depth filtering and depth-to-prior filtering. 
The prior-to-depth filtering transfers the accurate structural components from scene priors to depth. 
Conversely, the depth-to-prior filtering leverages depth knowledge to accentuate edges of these scene priors while downplaying the undesired textures. 

As a result, Fig. \ref{fig:FilterKernel}(g) and (h) show that the kernels produced by our SPFNet are largely resistant to interference and exhibit precise edges. 
Furthermore, Fig. \ref{fig:FilterKernel}(i) shows that the distribution of our normal and semantic filter kernels aligns more closely with the GT, as compared to DKN and DAGF.

%
The main contributions of this work are: 
\begin{itemize}
\item To address the issues of texture interference and edge inaccuracy in DSR, we are pioneering the incorporation of scene priors from large-scale models. 
\item We propose SPFNet, which recursively implements the novel all-in-one prior propagation, one-to-one prior embedding, and mutual guided filtering to further diminish texture interference and enhance edges. 
\item Extensive experiments on both real-world and synthetic datasets demonstrate that our SPFNet achieves superior performance, reaching the state-of-the-art. 
\end{itemize}

\section{Related Work}
\subsection{Multi-Modal Fusion based DSR}
Much progress in guided DSR~\cite{gu2017learning,zhong2021high,yan2022learning} based on deep learning has been made in recent years. 
Deng et al.~\cite{deng2020deep} utilize multi-modal convolutional sparse coding to extract the common features between RGB and depth. 
DCTNet~\cite{zhao2023spherical} develops spherical space feature decomposition to separate shared and private features.  
In~\cite{zhao2023spherical}, Zhao et al. project both RGB and depth features into a spherical space to separate their private features and align the shared ones.
Similarly, Yuan et al.~\cite{yuan2023recurrent} introduce a deep contrast network to split the depth into high-frequency and low-frequency maps.
Most recently, a few methods~\cite{wang2025multi,wang2026spatiotemporal,deng2023deepm} exploit depth structure recovery. 
For instance, DADA \cite{metzger2023guided} combines anisotropic diffusion and a convolutional network to improve the edge transferring property of diffusion. 
In \cite{wang2024sgnet}, SGNet designs a structure-guided network that employs the gradient and frequency domains for structure enhancement. To enhance the accuracy of DSR in real-world scenarios, He et al.~\cite{he2021towards} establish a real-world RGB-D benchmark and develop a baseline method for real-world DSR based on octave convolution.
DORNet~\cite{wang2025dornet} introduces a degradation-oriented and regularized framework, which selectively aggregates RGB and depth features by modeling the degradation representations between LR and HR depths.
More recently, some methods have attempted to introduce foundation models to enhance depth quality. For example, Yan et al.~\cite{yan2025ducos} utilize a depth estimation foundation model to provide dual constraints for HR depth restoration. Wang et al.~\cite{wang2025vggt} introduce a powerful depth-aware model on large-scale datasets, which directly predicts accurate depth from RGB. Viola et al.~\cite{viola2025marigold} use sparse depth as a condition for a pre-trained depth generation model to restore dense depth from RGB and sparse depth.
Unlike previous methods, which directly transform RGB features to depth, we focus more on leveraging normal and semantic priors to attenuate texture interference and improve accuracy near edges. 


\subsection{Guided Image Filtering based DSR}
To transfer the structure information from guidance to target, numerous guided image filtering methods ~\cite{he2012guided,shen2015mutual,zhong2023deep} have been proposed in recent years. 
Li et al.~\cite{li2019joint} develop joint image filtering based on deep convolutional networks to selectively transfer the structure from RGB to depth, and DKN~\cite{kim2021deformable} presents a deformable kernel network that explicitly generates a spatially-variant filter kernel and outputs sets of neighborhoods for each pixel. 
In~\cite{wang2023joint}, Wang et al. utilize the hybrid side window filtering to propagate multi-scale structure knowledge from RGB to depth. 
%
%
In contrast, our method focuses more on minimizing texture interference within filter kernels and improving edge accuracy by exploiting scene priors and similarity maps.

\subsection{Scene Prior Awareness}
Surface normal and semantic priors contain rich geometry and boundary information.
Recently, several methods~\cite{yang2018segstereo,qiu2019deeplidar,kirillov2023segment,wu2023learning,shao2024nddepth} have been designed to explore the use of these priors for facilitating downstream tasks.
Xu et al.~\cite{xu2019depth} learn the geometric constraints between depth and surface normal in a diffusion module to improve the performance of depth completion. Qiu et al.~\cite{qiu2019deeplidar} integrate guided image and surface normal to enhance the depth accuracy, and Fan et al.~\cite{fan2020sne} present a data-fusion CNN method that fuses inferred surface normal and image for accurate free space detection. 
In addition, SKF~\cite{wu2023learning} develops a semantic-aware knowledge-guided model to embed semantic prior in feature representation space for low-light image enhancement. In~\cite{jung2021fine}, Jung et al. incorporate semantics into geometric representation to improve self-supervised monocular depth estimation. 

Additionally, some methods attempt to leverage other prior to enhance the image. For instance, Xiao et al.~\cite{xiao2025spiking} optimize neuronal membrane potential and regulate spiking activity by designing a spiking attention block, while jointly modeling spatiotemporal features and mining non-local semantic priors to achieve efficient remote sensing image reconstruction. In~\cite{wang2025depth}, Wang et al. propose a coarse-to-fine fusion strategy that integrates accurate but incomplete metric depth with complete yet relative geometric structure priors, enabling zero-shot generalization for depth restoration.

In this work, we leverage large-scale models to generate accurate surface normal and semantic priors from HR RGB. Both are utilized as prompts to alleviate texture interference and facilitate the quality of depth restoration.

\subsection{Bidirectional Guidance}

The bidirectional guidance mechanism is widely applied in many computer vision tasks, achieving sufficient feature representation through information interaction. For example, Dong et al.~\cite{dong2022learning} design a cross-domain adaptive filter to achieve mutual modulation of multi-modal inputs, and DAGF~\cite{zhong2023deep} combines filter kernels from the guidance and target to model pixel-wise dependency between the two input images. In addition, Jiang et al.~\cite{jiang2025dawn} propose a direction-aware attention wavelet network based on mutual representation, which models the directionality of rain streaks via vector decomposition to achieve precise removal of heterogeneous rain streaks. Zhang et al.~\cite{zhang2024depth} introduce a dual-task collaborative mutual promotion framework, which achieves joint optimization of image dehazing and depth estimation through an alternating difference perception. Different from these approaches that use bidirectional guidance to fully fuse guidance and target features, our method focuses more on leveraging the structural correlation between scene priors and depth to mitigate texture interference.


 \begin{figure*}[t]
  \centering
  \includegraphics[width=0.94\textwidth]{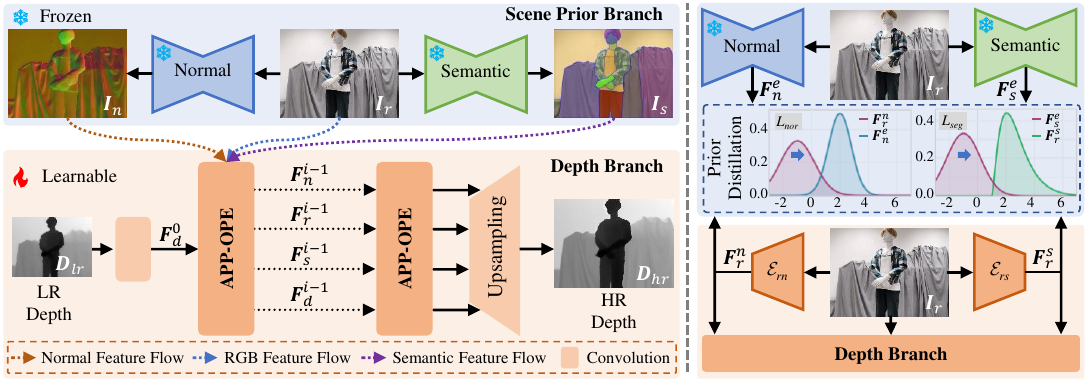}\\
  \caption{Overview of SPFNet (left) and its distilled variant SPFNet-D (right). For SPFNet, large-scale models are first employed to generate the normal prior $\boldsymbol I_{n}$ and semantic prior $\boldsymbol I_{s}$ from RGB $\boldsymbol I_{r}$. Then, $\boldsymbol I_{n}$ and $\boldsymbol I_{s}$ are fed together with $\boldsymbol I_{r}$ into multiple iteratively executed APP-OPE modules (APP and OPE are connected in series) to transfer scene prior knowledge into depth features, thereby reconstructing HR depth $\boldsymbol D_{hr}$. For SPFNet-D, the encoded features of normal and semantic models are separately extracted, yielding $\boldsymbol F_{n}^{e} $ and $\boldsymbol F_{s}^{e} $. Additionally, a prior distillation regularization (consisting of normal term $L _{nor} $ and semantic term $L _{sem} $) is introduced to distill normal and semantic features into RGB features $\boldsymbol F_{r}^{n} $ and $\boldsymbol F_{r}^{s} $, effectively reducing the computational overhead of large-scale models during inference. The distilled $\boldsymbol F_{r}^{n} $ and $\boldsymbol F_{r}^{s} $ replace the normal and semantic feature flows as input to the depth branch, whose architecture is kept identical to that of the original SPFNet.}\label{fig:pipeline}
\end{figure*}

\subsection{Image Super-Resolution Quality Assessment}

Recently, numerous image quality assessment methods have been proposed to automatically evaluate the quality of reconstructed images. Zhou et al.~\cite{zhou2022quality} introduce a super-resolution (SR) image fidelity index that adaptively combines deterministic fidelity and statistical fidelity through an uncertainty weighting scheme. Li et al.~\cite{li2024deep} design a full-reference bi-directional attention network for SR image quality assessment, enabling dynamic bidirectional interactions between the SR image generation process and the quality assessment process. In~\cite{zhou2020blind}, Zhou et al. develop a no-reference SR image quality assessment approach that adopts a two-stream convolutional network architecture to extract features related to structural degradation and texture distribution changes in SR images, thereby simulating the human visual system’s perception of image distortions. More recently, Li et al.~\cite{li2025perception} employ a bi-directional attention mechanism to model the bidirectional interaction between SR image generation and evaluation, integrating grouped multi-scale deformable convolution and sub-information excitation convolution to achieve adaptive assessment.

\section{Scene Prior Filtering Network}
In this section, we begin by introducing the overall architecture of our SPFNet and its distilled variant, termed SPFNet-D. We then provide a detailed description of the proposed all-in-one prior propagation (APP), one-to-one prior embedding (OPE), and mutual guided filtering (MGF). Finally, we present the loss function used for training.


\subsection{Network Architecture}
A naive guided DSR architecture generally incorporates an RGB guidance branch and a depth recovery branch. In our method, we introduce additional normal and semantic guidance branches, collectively forming the scene prior branch. Subsequently, in the depth recovery branch, the HR depth is progressively restored through multiple stages, under the guidance of the scene prior branch. 

As shown in Fig.~\ref{fig:pipeline}, our method comprises two variants: SPFNet (left) and SPFNet-D (right). Specifically, we first propose the original SPFNet, which leverages surface normal and semantic priors predicted by large-scale models as prompts to mitigate texture interference in RGB input while enhancing the representation of depth edges. To reduce the computational burden introduced by large-scale models during inference, we further introduce SPFNet-D. This variant distills the encoded features from surface normal and semantic models into RGB features to replace the direct use of estimated surface normal and semantic maps. Consequently, SPFNet-D effectively eliminates the additional computational overhead while attenuating texture interference in RGB features and refining depth edges.

\noindent \textbf{SPFNet.} As illustrated in Fig.~\ref{fig:pipeline} (left), SPFNet mainly consists of two components: the scene prior branch and depth branch. Given RGB $\boldsymbol I_{r} \in R^{sh\times sw\times 3}$ as input, the scene prior branch separately execute pre-trained large-scale models for surface normal and semantic prediction, obtaining the normal prior $\boldsymbol I_{n} \in R^{sh\times sw\times 3}$ and semantic prior $\boldsymbol I_{s} \in R^{sh\times sw\times 1}$. $h$ and $w$ are the height and width of LR depth, and $s$ is the upsampling factor. Subsequently, $\boldsymbol I_{r}$, $\boldsymbol I_{n}$, and $\boldsymbol I_{s}$ are each mapped into the feature space through $3\times3$ convolutional layers, generating the RGB feature flow $\boldsymbol F_{r}^{0}$, the normal feature flow $\boldsymbol F_{n}^{0}$, and the semantic feature flow $\boldsymbol F_{s}^{0}$.  In the depth branch, the LR depth $\boldsymbol D_{lr} \in R^{h\times w\times 1}$ is first encoded through convolutional layers into the initial depth features $\boldsymbol F_{d}^{0}$. These features, along with the RGB, normal, and semantic feature flows from the scene prior branch, are then fed into multiple APP-OPE modules (where APP and OPE are cascaded). Specifically, APP computes the structural similarity between all scene priors and the depth to filter out structural information in the scene priors that is highly correlated with the depth, thereby preliminarily reducing irrelevant texture interference. Building on this, OPE iteratively takes the scene priors refined by APP, together with the similarity weights, as input to generate filtering kernels without texture interference. This enables the transfer of knowledge from the large model that aligns with the similarity weights to the depth features. As a result, APP-OPE effectively suppress interference from RGB textures while enhancing depth representations, producing enhanced depth features $\boldsymbol F_{d}^{i}$, normal features $\boldsymbol F_{n}^{i}$, RGB features $\boldsymbol F_{r}^{i}$, and semantic features $\boldsymbol F_{s}^{i}$. Finally, an upsampling module (composed of a transposed convolution layer and a $3 \times 3$ convolution layer) is used to aggregate these  prior and depth features, thereby reconstructing HR depth $\boldsymbol D_{hr} \in R^{sh\times sw\times 1}$:
\begin{equation}
   \boldsymbol D_{hr} = f_{up} (\sqcup  ( \boldsymbol F_{n}^{i}, \boldsymbol F_{r}^{i}, \boldsymbol F_{s}^{i}, \boldsymbol F_{d}^{i} )) ,
\end{equation}
where $\sqcup (\cdot)$ represents concatenation operation.

 \begin{figure*}[t]
  \centering
  \includegraphics[width=0.91\textwidth]{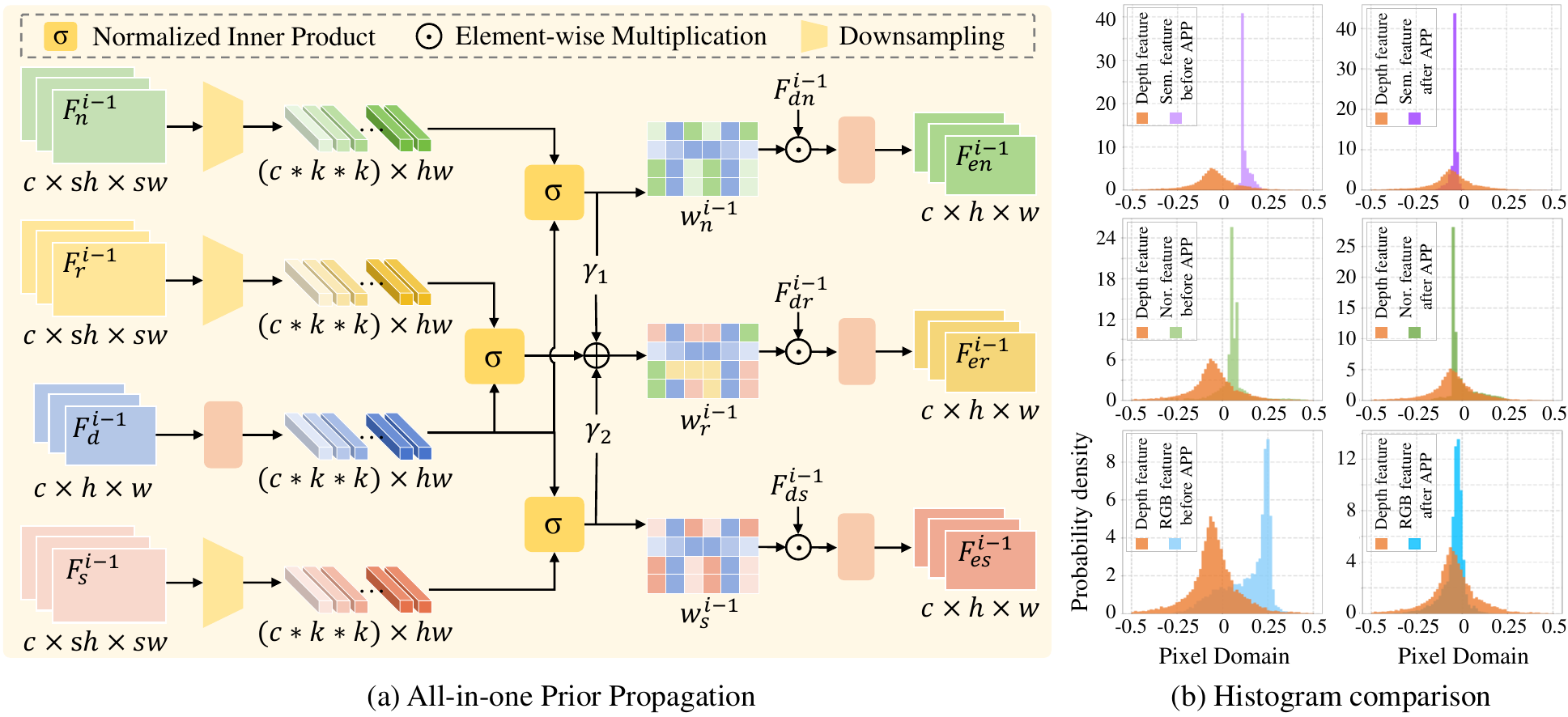}\\
  \caption{Scheme of (a) All-in-one Prior Propagation (APP), and (b) histogram comparison of scene prior features. Features $\boldsymbol F_{dn}^{i-1}$, $\boldsymbol F_{dr}^{i-1}$, and $\boldsymbol F_{ds}^{i-1} $ are downsampled from $\boldsymbol F_{n}^{i-1}$, $\boldsymbol F_{r}^{i-1}$, and $\boldsymbol F_{s}^{i-1} $, matching the size of depth features $\boldsymbol F_{d}^{i-1} $. The downsampling module consists of strided convolutions and ReLU activation functions.}\label{fig:APP}
\end{figure*}

To achieve a better balance between model complexity and reconstruction quality, we introduce a lightweight SPFNet-T. This model reduces the number of channels in all convolutional layers to one-seventh of those in SPFNet while keeping the original network architecture unchanged. As a result, the trainable parameter count of SPFNet-T is approximately $2.09\%$ of the original model.

\noindent \textbf{SPFNet-D.} Additionally, we further design a scene prior distillation variant, SPFNet-D, which effectively eliminates additional computational overhead from large-scale models. As shown on the right side of Fig. \ref{fig:pipeline}, our SPFNet-D first extracts encoded features from the normal model and the semantic model separately, obtaining $\boldsymbol F_{n}^{e}$ and $\boldsymbol F_{s}^{e}$. Then, we use encoders $\varepsilon  _{rn} $ and $\varepsilon  _{rs} $ to map the RGB input to features $\boldsymbol F_{r}^{n}$ and $\boldsymbol F_{r}^{s}$, where the shapes of $\boldsymbol F_{r}^{n}$ and $\boldsymbol F_{r}^{s}$ are consistent with $\boldsymbol F_{n}^{e}$ and $\boldsymbol F_{s}^{e}$, respectively. $\varepsilon _{rn}$ and $\varepsilon _{rs}$ first extract initial RGB features through multiple stacked $3\times3$ convolutional layers and ReLU activation layers. Then, multiple transposed convolutional layers and convolutional layers with a stride of 2 are composed to ensure the feature resolution is consistent with that of the large model encoder output.


Next, we introduce a prior distillation regularization to transfer normal and semantic knowledge from large-scale models into the RGB features, including a normal term $L_{nor} $ and a semantic term $L_{sem} $:
\begin{equation}\label{eq:distill}
\begin{split}
&L _{nor}\! =\! \sum_{k=1}^{C}\! \varphi\!\left ( \!\frac{\boldsymbol F_{N}^{n}}{\tau _{1} } \right ) \! \odot \! \left [ \!\Gamma \!\left ( \!\varphi \!\left ( \!\frac{\boldsymbol F_{N}^{n}}{\tau _{1} } \!\right )  \!\right ) \! - \!\Gamma\! \left (\! \psi \!\left ( \frac{\boldsymbol F_{N}^{nr}}{\tau _{1} } \!\right ) \!\right ) \! \right ] , \\
&L _{sem}\! =\!\sum_{k=1}^{C}\! \varphi\! \left ( \!\frac{\boldsymbol F_{N}^{s}}{\tau _{2} } \right ) \! \odot  \!\left [ \!\Gamma \!\left ( \!\varphi \!\left ( \!\frac{\boldsymbol F_{N}^{s}}{\tau _{2} } \!\right )  \!\right )\!  - \!\Gamma \!\left ( \!\psi \!\left ( \frac{\boldsymbol F_{N}^{sr}}{\tau _{2} } \!\right ) \!\right ) \! \right ] , 
\end{split}
\end{equation}
where $\boldsymbol F_{N}^{n}\!=\!f_{n} (\boldsymbol F_{n}^{e}(k))$, $\boldsymbol F_{N}^{nr}\!=\!f_{n} (\boldsymbol F_{r}^{n}(k))$, $\boldsymbol F_{N}^{s}\!=\!f_{n} (\boldsymbol F_{s}^{e}(k))$, and $\boldsymbol F_{N}^{sr}\!=\!f_{n} (\boldsymbol F_{r}^{s}(k))$. $C$ is the number of RGB feature channels, and $f_{n}$ represents normalization. $\Gamma (\cdot)$ denotes log function. $\varphi (\cdot)$ and $\psi (\cdot)$ are softmax and log softmax functions, respectively. $\tau _{1}$ and $\tau _{2}$ are temperature parameters, while $\odot $ refers to element-wise multiplication.

 \begin{figure*}[t]
  \centering
  \includegraphics[width=0.94\textwidth]{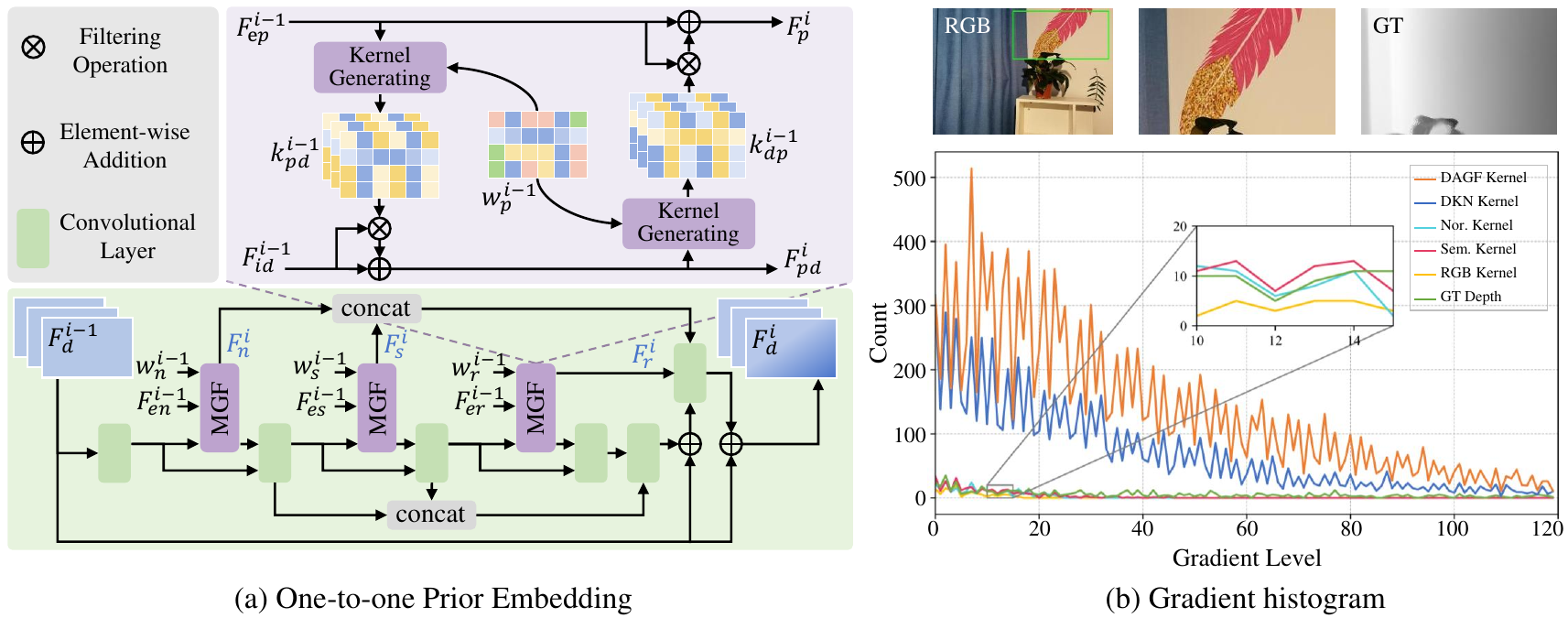}\\
  \caption{Scheme of (a) One-to-one Prior Embedding (OPE), and (b) gradient histogram of filter kernels in the texture area (green box). The surface normal, semantic, and RGB kernels are generated by our Mutual Guided Filtering (MGF).}\label{fig:OPE}
\end{figure*}

Finally, the distilled RGB features $\boldsymbol F_{r}^{n}$ and $\boldsymbol F_{r}^{s}$ are fed into the depth branch to replace the original normal and semantic feature flows of SPFNet. This replacement mitigates texture interference and enhances depth edge representations without introducing additional computational overhead. Furthermore, the depth branch in SPFNet-D is maintained identically to that of the original SPFNet.

\subsection{All-in-one Prior Propagation}
The proposed APP weakens the texture interference by using the inherent characteristics of scene priors from large-scale models. 
It computes the similarity between the multi-modal scene priors (RGB, normal, semantic, and depth), which is then used to mitigate the interference and promote the generation of prior filter kernels. 
As shown in Fig.~\ref{fig:APP}(a), we first downsample prior features $\boldsymbol F_{n}^{i-1}$, $\boldsymbol F_{r}^{i-1}$, and $\boldsymbol F_{s}^{i-1} $ to match the size of the depth features $\boldsymbol F_{d}^{i-1} $, producing downsampled features $\boldsymbol F_{dn}^{i-1}$, $\boldsymbol F_{dr}^{i-1}$, and $\boldsymbol F_{ds}^{i-1} $.

Next, APP unfolds (by $3\times 3$ kernel) depth features and downsampled scene prior features into patches, denoted as $n_{j}^{i-1}$, $r_{j}^{i-1}$, $s_{j}^{i-1}$, and $d_{j}^{i-1}$, where $j \in [1,h\times w]$. Subsequently, we compute the patch similarity $\sigma_{j}^{p}$ between the $j$-th prior patch $p_{j}^{i-1}$ (\emph{e.g.}, $r_{j}^{i-1}$, $n_{j}^{i-1}$, $s_{j}^{i-1}$) and depth patch $d_{j}^{i-1}$ using normalized inner product:
\begin{equation}\label{eq:nip}
 \sigma_{j}^{p}=\left \langle \frac{p_{j}^{i-1} }{\left \| p_{j}^{i-1} \right \| },\frac{d_{j}^{i-1} }{\left \| d_{j}^{i-1} \right \| }  \right \rangle. 
\end{equation}

The overall similarity between the scene prior and depth features can be derived by calculating the $\sigma_{j}^{p}$ of all patches, generating normal $\boldsymbol w_{n}^{i}$, semantic $\boldsymbol w_{s}^{i}$, and RGB $\boldsymbol w_{r}^{i}$ weights: 
\begin{equation}
\boldsymbol w_{r}^{i} =\gamma _{1} \circ  \boldsymbol w_{n}^{i} + \gamma _{2} \circ   \boldsymbol w_{s}^{i}+f_{r} (\sqcup _{j=1}^{h\times w}\sigma _{j}^{r} ),
\end{equation}
where $\boldsymbol w_{n}^{i} = f_{r} (\sqcup _{j=1}^{h\times w}\sigma _{j}^{n} )$, $\boldsymbol w_{s}^{i} = f_{r} (\sqcup _{j=1}^{h\times w}\sigma _{j}^{s} )$. $\sqcup (\cdot) $ and $\circ $ represent concatenation and scalar multiplication. $\gamma _{1}$ and $\gamma _{2}$ are learnable constant parameters initialized to zero, which are used to adaptively adjust the contribution of normal and semantic priors similarity to RGB. $f_{r}$ is a reshape function that transforms input to the dimensions of $\boldsymbol F_{dr}^{i-1}$.
 
Finally, the APP utilizes the similarity weights to attenuate interference in the scene prior features, generating enhanced normal features $\boldsymbol F_{en}^{i-1}$, RGB features $\boldsymbol F_{er}^{i-1}$, and semantic features $\boldsymbol F_{es}^{i-1}$. Fig.~\ref{fig:APP}(b) illustrates a comparison of the distributions between the prior features and depth features before and after using the APP module. It can be found that the output prior features are closer to the distribution of depth features, demonstrating that our APP succeeds in calibrating prior features and reducing interference.

\subsection{One-to-One Prior Embedding}
The OPE module continuously integrates each single-modal prior (normal, semantic, and RGB) into depth to further reduce interference and enhance edges. 
Fig.~\ref{fig:OPE}(a) shows the main steps of the OPE module. 
First, given the enhanced scene prior features ($\boldsymbol F_{en}^{i-1}$, $\boldsymbol F_{er}^{i-1}$, $\boldsymbol F_{es}^{i-1}$), similarity weights ($\boldsymbol w_{n}^{i-1}$, $\boldsymbol w_{r}^{i-1}$, $\boldsymbol w_{s}^{i-1}$), and depth features $\boldsymbol F_{d}^{i-1}$, OPE conducts MGF, denoted as $f_{m}$, to successively embed each single-modal prior into depth features:
\begin{equation}
\begin{split}
&\boldsymbol F_{n}^{i},\boldsymbol F_{nd}^{i} = f_{m} ( f_{c} ( \boldsymbol F_{d}^{i-1}  ),\boldsymbol F_{en}^{i-1},\boldsymbol w_{n}^{i-1}   ), \\
&\boldsymbol F_{s}^{i},\boldsymbol F_{sd}^{i} = f_{m} ( \boldsymbol F_{id1}^{i-1} ,\boldsymbol F_{es}^{i-1},\boldsymbol w_{s}^{i-1}   ), \\
&\boldsymbol F_{r}^{i},\boldsymbol F_{rd}^{i} = f_{m} (\boldsymbol F_{id2}^{i-1} ,\boldsymbol F_{er}^{i-1},\boldsymbol w_{r}^{i-1}   ),
\end{split}
\end{equation}
where $\boldsymbol F_{id1}^{i-1}\!=\!f_{c} ( \sqcup (f_{c} (\boldsymbol F_{d}^{i-1}) , \boldsymbol F_{nd}^{i}  ))$, $\boldsymbol F_{id2}^{i}\!=\!f_{c} (\sqcup( F_{id1}^{i}, \boldsymbol F_{ns}^{i} ) )$. $\boldsymbol F_{n}^{i}$, $\boldsymbol F_{s}^{i}$, and $\boldsymbol F_{r}^{i}$ indicate the filtered normal, semantic, and RGB features, respectively. 
In addition, $\boldsymbol F_{nd}^{i}$, $\boldsymbol F_{sd}^{i}$, and $\boldsymbol F_{rd}^{i}$ correspond to the filtered depth features of normal-to-depth, semantic-to-depth, and RGB-to-depth, respectively. $f_{c}$ denotes convolutional layer. 
Then, OPE aggregates the filtered scene prior and depth features:
\begin{equation}
   \boldsymbol F_{d}^{i}=f_{c}( \sqcup (\boldsymbol F_{dn}^{i},\boldsymbol F_{ds}^{i},\boldsymbol F_{dr}^{i},\boldsymbol F_{fd}^{i}))+ \boldsymbol F_{d}^{i-1} ,
\end{equation}
where $\boldsymbol F_{d}^{i}$ represents the enhanced depth features. $\boldsymbol F_{fd}^{i}=f_{c}( \sqcup (\boldsymbol F_{if1}^{i},\boldsymbol F_{if2}^{i},f_{c}(\boldsymbol F_{rd}^{i}, \boldsymbol F_{if2}^{i})))+ \boldsymbol F_{d}^{i-1} $.

\noindent \textbf{Mutual Guided Filtering.}
In contrast to existing guided filtering methods~\cite{kim2021deformable,dong2022learning,zhong2023deep,zhang2023bidirectional}, our MGF focuses more on employing scene priors and similarity weights to reduce texture interference within the filter kernel and enhance the accuracy of the edge representations. 
As depicted in the purple part of Fig.~\ref{fig:OPE}(a), it includes both prior-to-depth filtering (P2D) and depth-to-prior filtering (D2P).

For P2D, given the APP-enhanced scene prior features $\boldsymbol F_{ep}^{i-1}$ (\textit{e.g.}, $\boldsymbol F_{en}^{i-1}$, $\boldsymbol F_{er}^{i-1}$, $\boldsymbol F_{es}^{i-1}$), similarity weights $\boldsymbol w_{p}^{i-1}$ (\textit{e.g.}, $\boldsymbol w_{n}^{i-1} $, $\boldsymbol w_{r}^{i-1} $, $\boldsymbol w_{s}^{i-1} $), and depth features $\boldsymbol F_{id}^{i-1}$ as inputs, MGF first executes the kernel generator to construct a scene prior filtering kernel, denoted as $\boldsymbol k_{pd}^{i-1}$. The kernel generator consists of a $1\times1$ convolutional layer, a ReLU activation function layer, and a normalization operation. Furthermore, the prior-to-depth Filter Kernel Generator and the depth-to-prior Filter Kernel Generator share an identical network.

%
These kernels are subsequently applied to filter the depth features $\boldsymbol F_{id}^{i}$, thereby enabling the transfer of high-frequency components from the scene priors to the depth. 
The filtered depth features $\boldsymbol F_{pd}^{i}$ is described as:
\begin{equation}
\boldsymbol F_{pd}^{i}=\boldsymbol F_{id}^{i-1} \otimes \boldsymbol k_{pd}^{i-1}  + \boldsymbol F_{id}^{i-1},
\end{equation}
where $\boldsymbol k_{pd}^{i-1}=f_{kg}( \boldsymbol F_{ep}^{i-1}, \boldsymbol w_{p}^{i-1} )$. $f_{kg}$ stands for kernel generating module, consisting of a $1\times 1$ convolution and an activation function. $\otimes$ is the filtering operation.

Similarly, for depth-to-prior filtering, $\boldsymbol w_{p}^{i-1}$ and the filtered depth features $\boldsymbol F_{pd}^{i}$ are first fed into $f_{kg}$, generating the depth filter kernel $\boldsymbol k_{dp}^{i-1}$. 
Then, MGF filters prior features $\boldsymbol F_{ep}^{i-1}$ to preserve the structure required for the depth and further attenuate interference. 
The filtered prior features $\boldsymbol F_{p}^{i}$ is defined as:
\begin{equation}
\boldsymbol F_{p}^{i}=\boldsymbol F_{ep}^{i-1} \otimes \boldsymbol k_{dp}^{i-1}  + \boldsymbol F_{ep}^{i-1},
\end{equation}
where depth filter kernel $\boldsymbol k_{dp}^{i-1}=f_{kg}( \boldsymbol F_{pd}^{i}, \boldsymbol w_{p}^{i-1} )$. 

As illustrated in Fig.~\ref{fig:OPE}(b), we present the gradient histogram comparisons of filter kernels. Notably, the surface normal, semantic, and RGB kernels predicted by our method exhibit less gradient variations and are closer to the ground truth depth than other methods. These results further demonstrate that our MGF can efficiently mitigate texture interference and enhance edge representations.

\subsection{Loss Functions}
Given the predicted depth $\boldsymbol D_{hr}$ and ground-truth depth $\boldsymbol D_{gt}$ as input, we introduce a common reconstruction loss to  optimize our SPFNet:
\begin{equation}
   L_{total1} =\sum_{q\in Q}\left \| \boldsymbol D_{gt}^{q} -  \boldsymbol D_{hr}^{q} \right \|_{1},
\end{equation}
where Q is the set of valid pixels of $\boldsymbol D_{gt}$. $\left \| \cdot  \right \| _{1} $ is $L_{1}$ norm. 

Furthermore, to train the distillation variant SPFNet-D, we incorporate the reconstruction loss from original SPFNet with prior distillation regularization (as defined in Eq.~\eqref{eq:distill}):

\begin{equation}
   L_{total2} =\sum_{q\in Q}\left \| \boldsymbol D_{gt}^{q} -  \boldsymbol D_{hr}^{q} \right \|_{1} + \lambda _{1}L _{nor}+\lambda _{2}L _{sem},
\end{equation}
where $\lambda _{1}$ and $\lambda _{2}$ are hyperparameters, both set to $0.001$.

\begin{table*}[t]
\caption{Quantitative comparisons on synthetic DSR benchmarks. Following prior methods~\cite{he2021towards,zhao2022discrete}, RMSE in centimeters is used as the evaluation metric, with lower values indicating better performance. The \textbf{best} and \underline{second-best} results are marked.}\label{tab:Quantitative}
\centering
\renewcommand\arraystretch{1.05}
\resizebox{1\linewidth}{!}{
\begin{tabular}{lcccccccccccc|c}
\toprule 
\multirow{2}{*}{Methods}   &\multicolumn{3}{c}{NYU-v2}   &\multicolumn{3}{c}{RGB-D-D}  &\multicolumn{3}{c}{Lu}   &\multicolumn{3}{c|}{Middlebury} &\multirow{2}{*}{Venue}\\ 
\cmidrule(lr){2-4}\cmidrule(lr){5-7}\cmidrule(lr){8-10}\cmidrule(lr){11-13} 
 &$\times$4 &$\times$8 &$\times$16    &$\times$4 &$\times$8 &$\times$16 &$\times$4 &$\times$8 &$\times$16    &$\times$4 &$\times$8 &$\times$16 \\ \midrule
 \rowcolor[HTML]{E8E8E8}
\multicolumn{14}{c}{Bicubic downsampling}  \\ 
DJF~\cite{li2016deep}               &2.80 &5.33 &9.46       &3.41 &5.57 &8.15            &1.65 &3.96 &6.75                  &1.68 &3.24 &5.62           &ECCV 2016   \\
DSRNet~\cite{guo2018hierarchical}   &3.00 &5.16 &8.41       &-    &-    &-               &1.77 &3.10 &5.11                  &1.77 &3.05 &4.96           &TIP 2018   \\
DJFR~\cite{li2019joint}             &2.38 &4.94 &9.18       &3.35 &5.57 &7.99            &1.15 &3.57 &6.77                  &1.32 &3.19 &5.57           &PAMI 2019   \\
PAC~\cite{su2019pixel}              &1.89 &3.33 &6.78       &1.25 &1.98 &3.49            &1.20 &2.33 &5.19                  &1.32 &2.62 &4.58           &CVPR 2019   \\
CUNet~\cite{deng2020deep}           &1.92 &3.70 &6.78       &1.18 &1.95 &3.45            &0.91 &2.23 &4.99                  &1.10 &2.17 &4.33           &PAMI 2020   \\
DKN~\cite{kim2021deformable}        &1.62 &3.26 &6.51       &1.30 &1.96 &3.42            &0.96 &2.16 &5.11                  &1.23 &2.12 &4.24           &IJCV 2021   \\
FDKN~\cite{kim2021deformable}       &1.86 &3.58 &6.96       &1.18 &1.91 &3.41            &0.82 &2.10 &5.05                  &1.08 &2.17 &4.50           &IJCV 2021   \\
FDSR~\cite{he2021towards}           &1.61 &3.18 &5.86       &1.16 &1.82 &3.06            &1.29 &2.19 &5.00                  &1.13 &2.08 &4.39           &CVPR 2021   \\
GraphSR~\cite{de2022learning}       &1.79 &3.17 &6.02       &1.30 &1.83 &3.12            &0.92 &2.05 &5.15                  &1.11 &2.12 &4.43           &CVPR 2022   \\
SUFT~\cite{shi2022symmetric}        &1.12 &2.51 &4.86       &1.10 &\underline{1.69}&2.71 &1.10 &1.74 &3.92                  &1.07 &1.75 &3.18           &MM 2022   \\
DCTNet~\cite{zhao2022discrete}      &1.59 &3.16 &5.84       &\underline{1.08} &1.74&3.05 &0.88 &1.85 &4.39                  &1.10 &2.05 &4.19           &CVPR 2022  \\
DAGF~\cite{zhong2023deep}           &1.36 &2.87 &6.06       &1.14 &1.76 &2.82            &0.83 &1.93 &4.80                  &1.15 &1.80 &3.70           &TNNLS 2023   \\
RSAG~\cite{yuan2023recurrent}       &1.23 &2.51 &5.27       &1.14 &1.75 &2.96            &\textbf{0.79} &1.67 &4.30         &1.13 &2.74 &3.55           &AAAI 2023   \\
DADA~\cite{metzger2023guided}       &1.54 &2.74 &4.80       &1.20 &1.83 &2.80            &0.96 &1.87 &4.01                  &1.20 &2.03 &4.18           &CVPR 2023  \\
SSDNet~\cite{zhao2023spherical}     &1.60 &3.14 &5.86       &\textbf{1.04} &1.72 &2.92   &\underline{0.80} &1.82 &4.77      &\textbf{1.02} &1.91 &4.02  &ICCV 2023  \\
SGNet~\cite{wang2024sgnet}          &\underline{1.10} &2.44 &4.77   &1.10 &\textbf{1.64} &\underline{2.55}  
                                    &1.03 &1.61 &3.55      &1.15 &1.64 &2.95                            &AAAI 2024 \\
DORNet~\cite{wang2025dornet}        &1.19 &2.70 &5.60   &1.15 &1.80 &2.97  &0.92 &1.75 &4.41   &1.05 &1.76 &3.48                            &CVPR 2025 \\
\textbf{SPFNet-T}                   &1.52 &3.03 &5.71       &1.16 &1.77 &2.90      &\underline{0.80} &1.69 &4.31      &\underline{1.04} &1.77 &3.36     &IJCV 2026\\ 
\textbf{SPFNet-D}                   &\textbf{1.09} &\underline{2.39} &\underline{4.67}  &1.12 &1.71 &2.56  &0.92 &\underline{1.57} &\underline{3.22}   &1.06 &\underline{1.59} &\underline{2.87}                            &IJCV 2026\\
\textbf{SPFNet}                     &\textbf{1.09} &\textbf{2.36} &\textbf{4.55}   &1.13 &1.71 &\textbf{2.53}      
                                    &0.90 &\textbf{1.56} &\textbf{3.20}            &1.05 &\textbf{1.57} &\textbf{2.79}                                  &IJCV 2026\\
\midrule
\rowcolor[HTML]{E8E8E8}
\multicolumn{14}{c}{Nearest-neighbor downsampling}  \\  
DJF~\cite{li2016deep}               &3.54 &6.20 &10.21      &2.14 &3.32 &4.92           &2.54 &4.71 &7.66                   &2.14 &3.77 &6.12           &ECCV 2016  \\
DSRNet~\cite{guo2018hierarchical}   &3.49 &5.70 &9.76       &- &- &-                    &2.57 &4.46 &6.45                   &2.08 &3.26 &5.78           &TIP 2018  \\
DJFR~\cite{li2019joint}             &3.38 &5.86 &10.11      &1.90 &3.11 &4.89           &2.22 &4.54 &7.48                   &1.98 &3.61 &6.07           &PAMI 2019  \\
PAC~\cite{su2019pixel}              &2.82 &5.01 &8.64       &- &- &-                    &2.48 &4.37 &6.60                   &1.91 &3.20 &5.60           &CVPR 2019  \\
CUNet~\cite{deng2020deep}           &4.09 &6.25 &10.23      &1.85 &3.07 &5.01           &2.15 &4.33 &7.72                   &2.06 &3.97 &6.36           &PAMI 2020  \\
DKN~\cite{kim2021deformable}        &2.46 &4.76 &8.50       &1.92 &2.91 &4.46           &2.35 &4.16 &6.33                   &1.93 &3.17 &5.49           &IJCV 2021  \\
FDKN~\cite{kim2021deformable}       &2.62 &4.99 &8.67       &1.84 &2.93 &4.76           &2.64 &4.55 &7.20                   &2.21 &3.64 &6.15           &IJCV 2021  \\
FDSR~\cite{he2021towards}           &2.50 &4.62 &7.77       &1.83 &2.77 &4.07           &2.17 &3.97 &6.51                   &1.85    &2.97 &5.31           &CVPR 2021 \\
DCTNet~\cite{zhao2022discrete}      &2.56 &4.89 &9.11       &1.86 &2.86 &4.28           &2.31 &4.17 &6.69                   &1.81 &2.96 &5.39           &CVPR 2022 \\
SUFT~\cite{shi2022symmetric}        &2.05 &4.11 &7.26       &1.85 &2.79 &3.95      
                                    &2.07 &\underline{3.63} &6.16                &1.76 &2.76 &5.16                              &MM 2022  \\
DAGF~\cite{zhong2023deep}           &2.35 &4.62 &7.81                                        &\underline{1.78} &\textbf{2.65} &3.95      
                                    &\underline{1.96} &3.81 &6.16                            &1.78 &2.73 &4.75      &TNNLS 2023     \\
RSAG~\cite{yuan2023recurrent}       &2.67 &4.73 &7.66       &1.80 &2.79 &4.01           &2.11 &3.91 &6.32                   &1.73 &2.98 &5.05           &AAAI 2023   \\
SGNet~\cite{wang2024sgnet}          &\underline{1.94} &4.04 &7.31   &1.81 &\underline{2.66} &\textbf{3.74}  &1.98 &3.58 &5.82   &1.72 &2.76 &\underline{4.43}                            &AAAI 2024 \\
DORNet~\cite{wang2025dornet}        &2.02 &\underline{4.10} &7.55   &1.89 &2.85 &4.03  &2.08 &3.72 &6.14   &1.73 &2.72 &4.68                            &CVPR 2025 \\
\textbf{SPFNet-T}                   &2.53 &4.83 &8.24       &1.88 &2.90 &4.17           &2.35 &4.26 &6.66                   &1.77 &2.96 &4.91           &IJCV 2026\\ 
\textbf{SPFNet-D}                   &\textbf{1.92} &\textbf{3.96} &\underline{7.10}   &1.84 &2.74 &\underline{3.83}  &2.07 &3.65 &\underline{5.78}   &\underline{1.71} &\underline{2.64} &\textbf{4.39}     &IJCV 2026\\
\textbf{SPFNet}                     &2.24 &\textbf{3.96} &\textbf{6.74}          &\textbf{1.76} &2.76 &\textbf{3.74}      
                                    &\textbf{1.94} &\textbf{3.46} &\textbf{5.48}          &\textbf{1.65} &\textbf{2.62} &\textbf{4.39}                  &IJCV 2026\\   
\bottomrule
\end{tabular}}
\end{table*}

\section{Experimental Results}
In this section, we conduct extensive experiments to evaluate the performance of our proposed SPFNet. Specifically, we first elaborate on the experimental setup and implementation details. Then, we present the comparison results with previous state-of-the-art methods across several experiments: synthetic DSR, real-world DSR, model complexity analysis, and joint DSR with denoising. To further demonstrate the generalization capability of our approach, we also conduct experiments on other multi-modal image restoration tasks. Finally, we perform multiple ablation studies to comprehensively evaluate the effectiveness of our approach. 

\subsection{Experimental Setups and Implementation Details}
We carry out extensive experiments on five DSR datasets, including NYU-v2~\cite{silberman2012indoor}, Lu~\cite{lu2014depth}, Middlebury~\cite{hirschmuller2007evaluation,scharstein2007learning}, RGB-D-D~\cite{he2021towards}, and TOFDSR~\cite{yan2025tri}. We utilize the Root Mean Square Error (RMSE) metric in centimeters to evaluate the DSR methods~\cite{kim2021deformable,shi2022symmetric,zhao2023spherical}.
During training, the scene priors and GT depth are cropped to $256 \times 256$. We employ the Adam~\cite{kingma2014adam} optimizer with an initial learning rate of $1 \times 10^{-4}$ to train our method. The proposed model is implemented in PyTorch with a single NVIDIA RTX 4090.

\subsection{Comparison with the State-of-the-Art DSR methods}
We compare our method with state-of-the-art approaches on $\times 4$, $\times 8$, $\times 16$, and $\times 32$ DSR, including DJF \cite{li2016deep}, DSRNet \cite{guo2018hierarchical}, DJFR \cite{li2019joint}, PAC \cite{su2019pixel}, CUNet \cite{deng2020deep}, DKN \cite{kim2021deformable}, FDKN \cite{kim2021deformable}, FDSR~\cite{he2021towards}, SUFT \cite{shi2022symmetric}, DCTNet \cite{zhao2022discrete}, GraphSR \cite{de2022learning}, DAGF \cite{zhong2023deep}, RSAG \cite{yuan2023recurrent}, DADA \cite{metzger2023guided}, SSDNet \cite{zhao2023spherical}, SGNet \cite{wang2024sgnet}, and DORNet \cite{wang2025dornet}.

\begin{table*}[t]
\caption{Quantitative comparisons at large-scale factor ($\times 32$) using bicubic downsampling.}\label{tab:X32_64}
	\centering
	\LARGE
	\resizebox{1\linewidth}{!}{
\begin{tabular}{c|cccccccccc}
\toprule 
Datasets  &DJFR~\cite{li2019joint} &CUNet~\cite{deng2020deep} & DKN~\cite{kim2021deformable}   &FDSR~\cite{he2021towards}   &DCTNet~\cite{zhao2022discrete}   	&SGNet~\cite{wang2024sgnet}	&DORNet~\cite{wang2025dornet} &\textbf{SPFNet-T}&\textbf{SPFNet-D}&\textbf{SPFNet}\\
\midrule
Lu              &9.94     &10.97    &8.98         &8.62     &9.28       &7.23     &8.28     &7.32    &\underline{6.42}     &\textbf{6.39} 	\\
NYU-v2           &14.12    &15.95    &12.46      &14.19    &13.33      &11.24    &10.97    &10.07   &\underline{8.69}     &\textbf{8.06} 	\\
RGB-D-D    	    &6.48     &7.75     &5.97         &5.08     &5.99       &5.15     &5.05     &4.80    &\underline{4.13}     &\textbf{3.97}  \\
Middlebury       &8.57    &9.23    &7.76      &7.26    &8.22      &6.79    &6.87    &6.09   &\textbf{5.59}     &\underline{5.90} 	\\
\bottomrule
\end{tabular}}
\end{table*}

\begin{figure*}[t]
\centering
\includegraphics[width=1\linewidth]{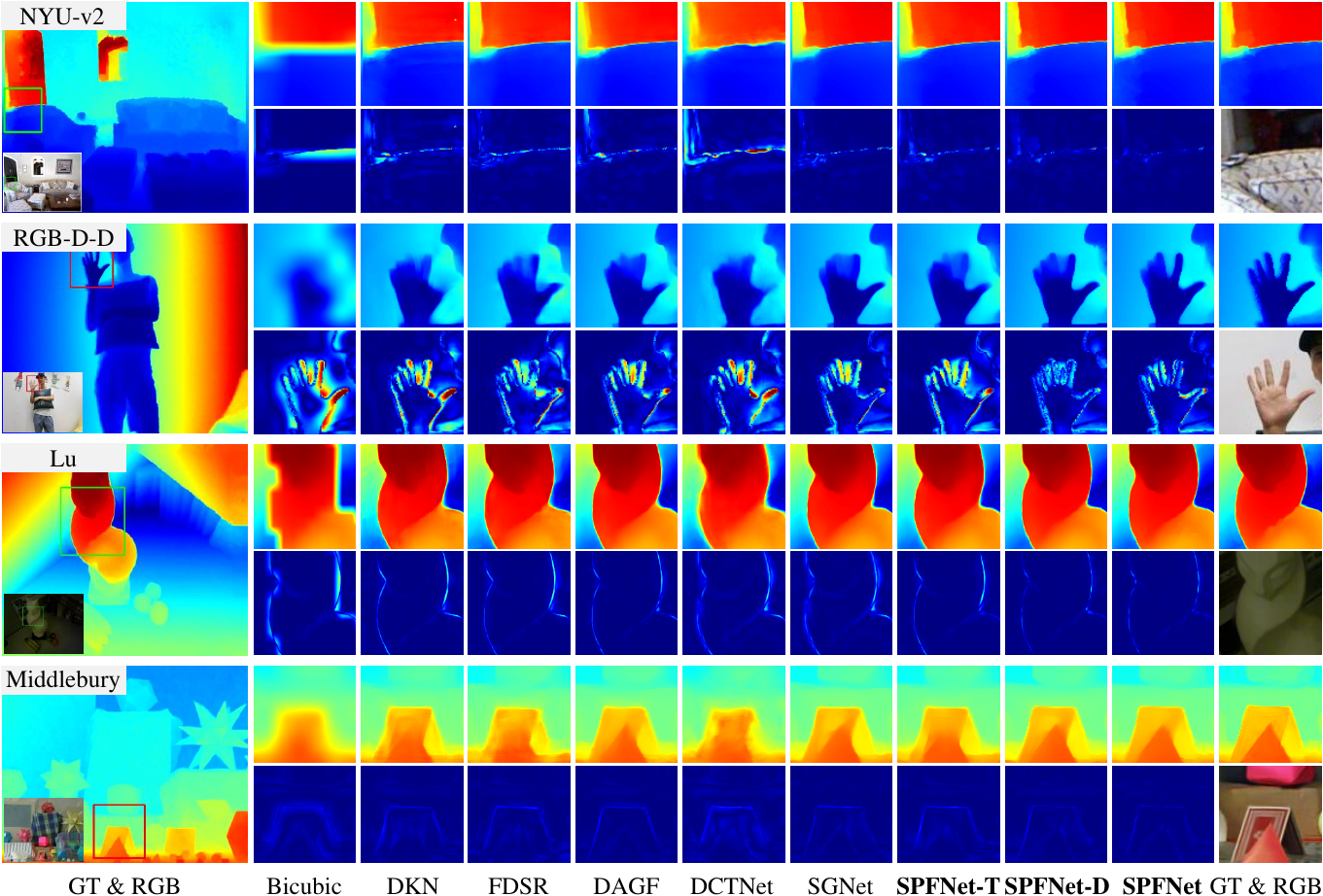}
\caption{Visual results and error maps on four synthetic datasets ($\times16$). Brighter colors in error maps indicate larger errors.}
\label{fig:NRLM_X16}
\end{figure*}

\begin{figure*}[!ht]
\centering
\includegraphics[width=1\linewidth]{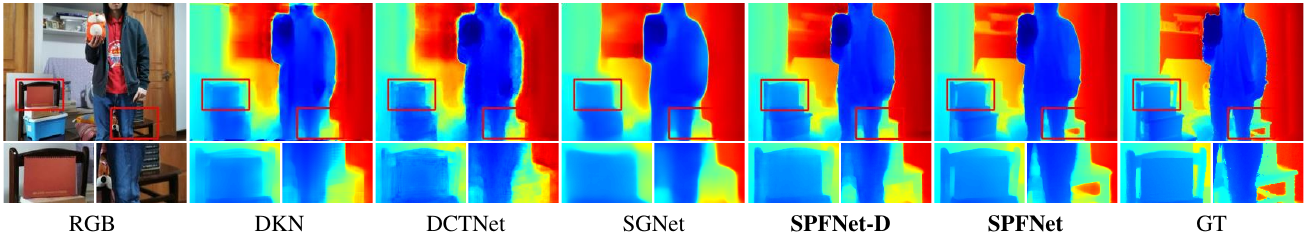}
\caption{Visual results on the synthetic RGB-D-D dataset ($\times32$).}
\label{fig:RGBDD_x32}
\end{figure*}

\noindent \textbf{Experimental Results on the Synthetic DSR.}
To assess the performance of our method on synthetic datasets, we conduct extensive experiments on the NYU-v2, Lu, Middlebury, and RGB-D-D datasets. Similar to previous  methods~\cite{kim2021deformable,zhao2022discrete,zhong2023deep,zhao2023spherical}, the LR depth is produced by downsampling the GT depth using bicubic interpolation and nearest-neighbor interpolation, respectively. Our model is trained on the $1,000$ RGB-D pairs of NYU-v2, with the remaining $449$ pairs for testing. Furthermore, the model pre-trained on the NYU-v2 dataset is applied directly to evaluate the Lu ($6$ RGB-D pairs), Middlebury ($30$ pairs), and RGB-D-D ($405$ pairs) datasets without any fine-tuning. 

Tab.~\ref{tab:Quantitative} demonstrates that our SPFNet achieves state-of-the-art performance across multiple synthetic datasets under different downsampling (bicubic and nearest-neighbor). Specifically, as shown in Tab. \ref{tab:Quantitative}, SPFNet surpasses most competing methods on four benchmark datasets under various downsampling. For example, for bicubic downsampling, our method decreases the RMSE ($\times 16$) by $0.22cm$ on NYU-v2 and by $0.35cm$ on Lu compared to the suboptimal approach. For nearest-neighbor downsampling, SPFNet outperforms the second-best method by $0.52cm$ on $\times 16$ NYU-v2 and $0.34cm$ on $\times 16$ Lu. Additionally, our SPFNet-D also delivers competitive accuracy while maintaining satisfactory efficiency. Specifically, SPFNet-D exceeds the second-best approach with RMSE reductions of $0.10cm$ under bicubic downsampling and $0.16cm$ under nearest-neighbor downsampling on the $\times16$ NYU-v2 dataset.

Fig. \ref{fig:NRLM_X16} presents the visual results on synthetic datasets with bicubic downsampling.  It is evident that our method predicts depth with superior edge accuracy. For example, the edges of the human hand and the sculpture are more distinct than others, and the error maps display smaller errors. 

Furthermore, Tab.~\ref{tab:X32_64} lists the quantitative comparisons on $\times 32$ DSR, showing that both our SPFNet and SPFNet-D achieve the best performance at this large scale factor. For instance, compared to the second-best approach, our SPFNet reduces the RMSE by $2.01cm$ on the NYU-v2 dataset and by $1.08cm$ on the RGB-D-D dataset.

Furthermore, Fig. \ref{fig:RGBDD_x32} presents a visual comparison on $\times32$ RGB-D-D dataset, demonstrating that our method maintains satisfactory performance even at large scale factors. For instance, the structure and edges of the chair predicted by our SPFNet and SPFNet-D in Fig. \ref{fig:RGBDD_x32} are more accurate and closer to the ground truth compared to other methods.

\begin{table*}[t]
\caption{Quantitative comparisons with existing state-of-the-art methods on the real-world RGB-D-D and TOFDSR datasets.}\label{tab:Real}
	\centering
	\LARGE
	\resizebox{1\linewidth}{!}{
\begin{tabular}{c|cccccccccccc}
\toprule 
Datasets	 &DJFR~\cite{li2019joint} &CUNet~\cite{deng2020deep} & DKN~\cite{kim2021deformable}    & FDSR~\cite{he2021towards}   &DCTNet~\cite{zhao2022discrete}   &SUFT~\cite{shi2022symmetric}	  &SSDNet~\cite{zhao2023spherical} 	&SGNet~\cite{wang2024sgnet}	&\textbf{SPFNet-T}&\textbf{SPFNet-D}&\textbf{SPFNet}\\
\midrule
RGB-D-D          &5.52    &5.84    &5.08     &5.49    &5.43    &5.41      &5.38      &5.32             &4.68   &\textbf{3.71}     & \underline{4.21}	\\
TOFDSR     	     &5.72    &6.04    &5.50      &5.03    &5.16 	 &4.37     &-        &\textbf{4.33} &5.13   &\underline{4.51}        &4.58 \\
\bottomrule
\end{tabular}}
\end{table*}

\begin{figure*}[t]
\centering
\includegraphics[width=1\linewidth]{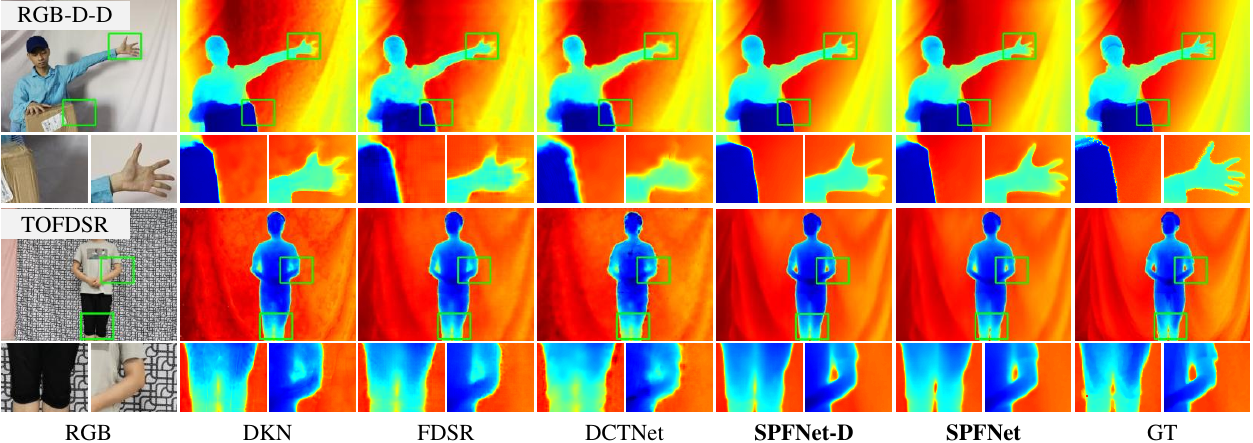}
\caption{Visual results on the real-world RGB-D-D and TOFDSR datasets.}
\label{fig:RGBDD_Real}
\end{figure*}

 \begin{figure*}[!ht]
  \centering
  \includegraphics[width=1\textwidth]{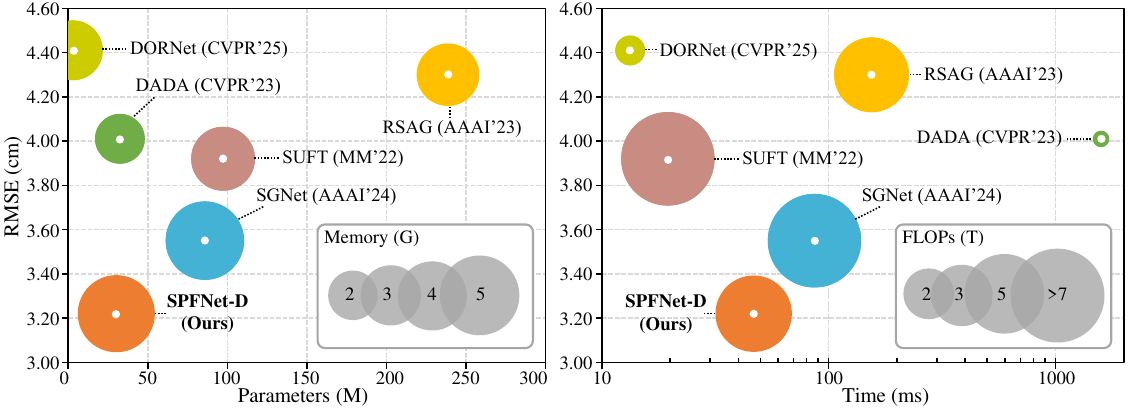}\\
  \caption{Model complexity comparisons on the $\times16$ Lu dataset (bicubic downsampling), where inference time is tested on a single 4090 GPU. Left: Parameters and Memory comparison. Right: Inference time and FLOPs comparison.}\label{fig:Params_Time_X16Lu}
\end{figure*}

\noindent \textbf{Experimental Results on the Real-world DSR.}
We perform experiments on real-world RGB-D-D and TOFDSR datasets to evaluate the performance of DSR methods in real environments, where LR and ground-truth depth are captured in real-world scenes. Specifically, RGB-D-D consists of $2,215$ RGB-D pairs for training and $405$ pairs for testing. Additionally, we employ the colorization approach \cite{levin2004colorization} to fill in the raw LR depth in TOFDC \cite{yan2024tri} as new LR depth input, resulting in the TOFDSR dataset, which includes $10K$ RGB-D pairs in the training set and $560$ pairs in the test set.


As shown in Tab.~\ref{tab:Real}, our method achieves the best performance on the RGB-D-D dataset and delivers competitive results on the TOFDSR dataset.  For example, SPFNet-D reduces RMSE by $30.26\%$ compared to the second-best SGNet on RGB-D-D. Unlike synthetic datasets, LR depth captured in real-world environments often exhibits severe structural distortion. Fig.~\ref{fig:RGBDD_Real} shows the visual results on the real-world RGB-D-D and TOFDSR, demonstrating that our approach recovers more accurate edges compared to others. For instance, the edges of the hand (RGB-D-D) and the arm (TOFDSR) predicted by SPFNet in Fig.~\ref{fig:RGBDD_Real} are more closely aligned with the GT depth. Overall, these quantitative and visual results indicate that our method can effectively improve DSR performance in real-world scenarios.

\begin{table*}[t]
\caption{Quantitative comparisons of joint DSR and denoising on four benchmark datasets.}\label{tab:denoising_NYU}
\centering
\renewcommand\arraystretch{1.05}
\resizebox{0.88\linewidth}{!}{
\begin{tabular}{lcccccccccccc}
\toprule 
\multirow{2}{*}{Methods}   &\multicolumn{3}{c}{NYU-v2}   &\multicolumn{3}{c}{RGB-D-D}  &\multicolumn{3}{c}{Lu}   &\multicolumn{3}{c}{Middlebury} \\ 
\cmidrule(lr){2-4}\cmidrule(lr){5-7}\cmidrule(lr){8-10}\cmidrule(lr){11-13} 
 &$\times$4 &$\times$8 &$\times$16    &$\times$4 &$\times$8 &$\times$16 &$\times$4 &$\times$8 &$\times$16    &$\times$4 &$\times$8 &$\times$16 \\ \midrule
 \rowcolor[HTML]{E8E8E8}
\multicolumn{13}{c}{Gaussian noise to LR depth}  \\
DJF~\cite{li2016deep}               &8.53 &12.38 &17.86      &3.06 &4.37 &6.44            &4.87 &7.43 &11.25                 &5.31 &7.69 &10.63 \\
DJFR~\cite{li2019joint}             &7.99 &11.65 &17.06      &2.86 &4.16 &6.11            &4.40 &6.98 &10.65                 &5.01 &7.28 &10.36 \\
CUNet~\cite{deng2020deep}           &7.80 &10.72 &15.52      &2.74 &3.98 &5.76            &4.07 &6.54 &9.84                  &4.77 &6.81 &9.80  \\
DKN~\cite{kim2021deformable}        &7.06 &10.26 &15.40      &2.72 &3.92 &5.89            &4.21 &6.36 &9.81                  &4.33 &6.67 &9.77  \\
FDKN~\cite{kim2021deformable}       &7.41 &10.69 &15.72      &2.73 &4.03 &6.09            &4.12 &6.54 &11.63                 &4.48 &6.66 &9.86  \\
FDSR~\cite{he2021towards}           &5.99 &8.68 &13.28       &2.44 &3.48 &4.99            &3.77 &6.24 &9.87                  &-    &5.73 &8.66  \\
SUFT~\cite{shi2022symmetric}        &5.46 &7.94 &12.23       &2.46 &3.42 &4.85            &3.59 &6.01 &9.90      &3.51 &5.61 &8.70        \\
DCTNet~\cite{zhao2022discrete}      &6.41 &9.81 &14.81       &2.55 &3.68 &5.30            &3.91 &6.27 &9.56                  &3.87 &6.06 &8.89  \\
SGNet~\cite{wang2024sgnet}          &5.33 &7.95 &\underline{12.19}       &2.41 &\underline{3.30} &\underline{4.61}            &\textbf{3.29} &5.45 &\underline{8.27}  &3.33 &5.11 &7.88   \\
DORNet~\cite{wang2025dornet}          &5.69 &8.65 &13.70   &2.53 &3.55 &5.21  &3.69 &6.08 &9.34   &3.55 &5.10 &7.72                            \\
\textbf{SPFNet-T}                   &5.94 &8.48 &12.56       &2.47 &3.43 &4.79            &3.60 &5.69 &8.43                  &3.49 &5.26 &\underline{7.68}        \\
\textbf{SPFNet-D}          &\underline{5.26} &\underline{7.86} &12.32   &\underline{2.37} &3.32 &4.71  &3.39 &\underline{5.39} &8.38   &\underline{3.30} &\underline{4.99} &7.69                            \\
\textbf{SPFNet}                     &\textbf{5.14} &\textbf{7.44} &\textbf{10.96}   &\textbf{2.36} &\textbf{3.29} &\textbf{4.47}      
                                    &\underline{3.34} &\textbf{5.30} &\textbf{7.93}             &\textbf{3.23} &\textbf{4.85} &\textbf{7.20}          \\
\midrule
\rowcolor[HTML]{E8E8E8}
\multicolumn{13}{c}{Gaussian noise to both LR depth and RGB image}  \\
DJF~\cite{li2016deep}               &8.91 &13.14 &18.72      &3.13 &4.56 &6.81            &4.98 &7.49 &11.69                 &5.43 &7.82 &10.89       \\
DJFR~\cite{li2019joint}             &8.47 &12.48 &18.11      &2.90 &4.32 &6.56            &4.46 &7.07 &10.97                 &5.14 &7.50 &10.64       \\
CUNet~\cite{deng2020deep}           &8.32 &11.49 &16.60      &2.90 &4.04 &6.07            &4.39 &6.43 &10.11                 &5.12 &6.97 &9.90        \\
DKN~\cite{kim2021deformable}        &7.54 &11.04 &16.77      &2.74 &4.05 &6.34            &4.09 &6.27 &10.39                 &4.48 &6.88 &9.99        \\
FDKN~\cite{kim2021deformable}       &7.54 &11.23 &16.92      &2.76 &4.12 &6.37            &4.09 &6.53 &10.59                 &4.50 &6.94 &10.12       \\
FDSR~\cite{he2021towards}           &6.88 &10.30 &16.00      &2.47 &3.70 &5.79            &3.69 &6.20 &10.21                 &- &6.30 &9.33           \\
SUFT~\cite{shi2022symmetric}        &6.65 &9.93 &15.55       &2.39 &3.59 &5.64            &3.49 &5.90 &10.08                 &3.75 &6.12 &9.23        \\
DCTNet~\cite{zhao2022discrete}      &7.38 &10.99 &16.90      &2.56 &3.87 &6.05            &3.81 &6.18 &10.16                 &4.22 &6.65 &9.75        \\
SGNet~\cite{wang2024sgnet}          &6.22 &9.92 &15.50       &\underline{2.37} &3.55 &5.58            &3.44 &\underline{5.58} &9.41                  &3.61 &5.92 &8.99        \\
DORNet~\cite{wang2025dornet}        &6.03 &8.69 &13.32   &2.47 &3.49 &5.09  &3.68 &6.01 &9.41   &3.70 &5.47 &\underline{7.82}                            \\
\textbf{SPFNet-T}                   &\underline{6.04} &\underline{8.60} &\underline{12.67}   &2.42 &\underline{3.44} &\underline{4.77}            
                                    &3.57 &5.69 &\underline{8.76}                            &\underline{3.54} &\underline{5.28} &7.93        \\
\textbf{SPFNet-D}                   &6.59 &9.79 &15.45   &2.40 &3.61 &5.67  &\underline{3.33} &5.64 &9.34   &3.66 &5.94 &8.90                            \\
\textbf{SPFNet}                     &\textbf{5.34} &\textbf{7.66} &\textbf{11.31}       &\textbf{2.34} &\textbf{3.28} &\textbf{4.48}            
                                    &\textbf{3.31} &\textbf{5.50} &\textbf{8.22}        &\textbf{3.29} &\textbf{5.05} &\textbf{7.73}        \\
\bottomrule
\end{tabular}}
\end{table*}

\begin{figure*}[!ht]
\centering
\includegraphics[width=1\linewidth]{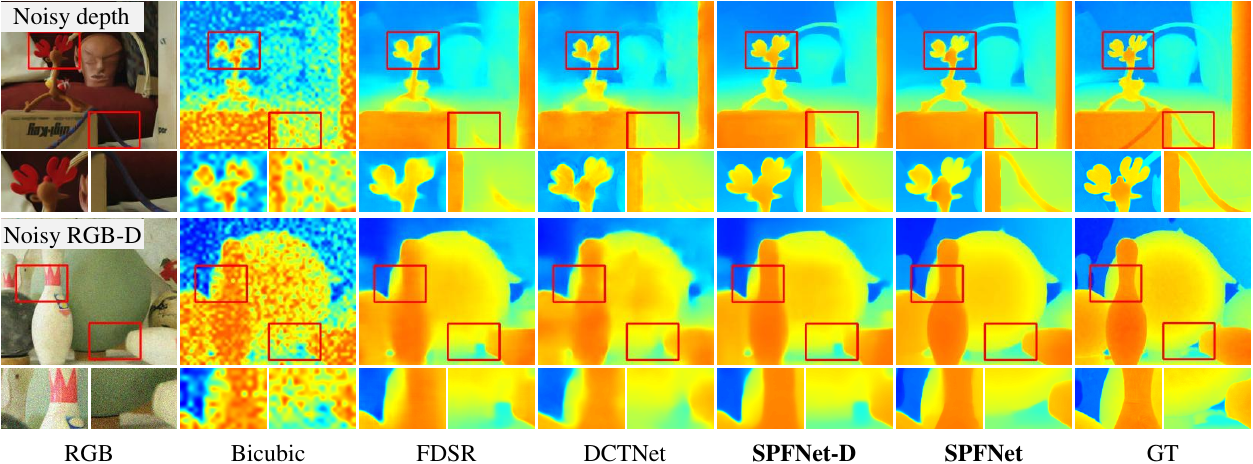}
\caption{Visual results of joint DSR and denoising on the $\times 8$ Middlebury dataset.}
\label{fig:Denoising}
\end{figure*}

\begin{table*}[t]
\caption{Quantitative comparisons of Pan-Sharpening on WorldView III and GaoFen2 datasets.}\label{tab:pan}
\centering
\small
\renewcommand\arraystretch{1.05}
\resizebox{0.98\linewidth}{!}{
\begin{tabular}{l|cccc|cccc}
\toprule
 \multirow{2}{*}{Methods} &\multicolumn{4}{c|}{WorldView III} &\multicolumn{4}{c}{GaoFen2} \\
 \cmidrule{2-9}
         &PSNR$\uparrow$ &SSIM$\uparrow$ &SAM\cite{yuhas1992discrimination}$\downarrow$ &ERGAS~\cite{alparone2007comparison}$\downarrow$        
         &PSNR$\uparrow$ &SSIM$\uparrow$ &SAM\cite{yuhas1992discrimination}$\downarrow$  &ERGAS~\cite{alparone2007comparison}$\downarrow$\\ \midrule

GFPCA~\cite{liao2015two}           &22.3344 &0.4826 &0.1294 &8.3964                                                 &37.9443 &0.9204 &0.0314 &1.5604   \\
PanNet~\cite{yang2017pannet}       &29.6840 &0.9072 &0.0851 &3.4263                                                 &43.0659 &0.9685 &0.0178 &0.8577    \\
MSDCNN~\cite{yuan2018multiscale}   &30.3038 &0.9184 &0.0782 &3.1884                                                 &45.6874 &0.9827 &0.0135 &0.6389   \\
SRPPNN~\cite{cai2020super}         &30.4346 &0.9202 &0.0770 &3.1553                                                 &47.1998 &0.9877 &0.0106 &0.5586   \\
GPPNN~\cite{xu2021deep}            &30.1785 &0.9175 &0.0776 &3.2593                                                 &44.2145 &0.9815 &0.0137 &0.7361   \\
MutInf~\cite{zhou2022mutual}       &\underline{30.4907} &\underline{0.9223} &\textbf{0.0749} &\underline{3.1125}    &\underline{47.3042} &\underline{0.9892} &\underline{0.0102} &\underline{0.5481}   \\
PanFlow~\cite{yang2023panflownet}  &30.4873 &0.9221 &\underline{0.0751} &3.1142                                     &47.2533 &0.9884 &0.0103 &0.5512   \\
\textbf{SPFNet}&\textbf{30.6786}&\textbf{0.9244}&0.0769&\textbf{3.0447}                                             &\textbf{47.4818}&\textbf{0.9895}&\textbf{0.0101}&\textbf{0.5441}   \\
\bottomrule
\end{tabular}}
\end{table*}

\begin{table*}[t]
\caption{Quantitative comparisons of saliency map super-resolution on DUT-OMRON~\cite{yang2013saliency}. The metric is Fscore.}\label{tab:saliencysr}
\Large
	\centering
	\resizebox{1\linewidth}{!}{
\begin{tabular}{c|cccccccccc}
\toprule 
Fscore		&DJFR~\cite{li2019joint} &CUNet~\cite{deng2020deep} & DKN~\cite{kim2021deformable}  & FDKN~\cite{kim2021deformable}  & FDSR~\cite{he2021towards}   & SUFT~\cite{shi2022symmetric}   &DCTNet~\cite{zhao2022discrete}  &SGNet~\cite{wang2024sgnet} 	&\textbf{SPFNet}\\
\midrule
$\times 4$ 	  &0.9858    &0.9863    &0.9917    &0.9931    &-         &0.9948   &0.9874  &\underline{0.9951}  &\textbf{0.9967}   \\
$\times 8$    &0.9599    &0.9497    &0.9389    &0.9789    &0.9819    &0.9825    &0.9619  &\underline{0.9829}    &\textbf{0.9844}   \\
$\times 16$   &0.8975    &0.8972    &\underline{0.9566}    &0.9543    &0.9549    &\textbf{0.9584}    &0.9003 &0.9483  &0.9475   \\
\bottomrule
\end{tabular}}
\end{table*}

\begin{figure*}[!ht]
\centering
\includegraphics[width=1\linewidth]{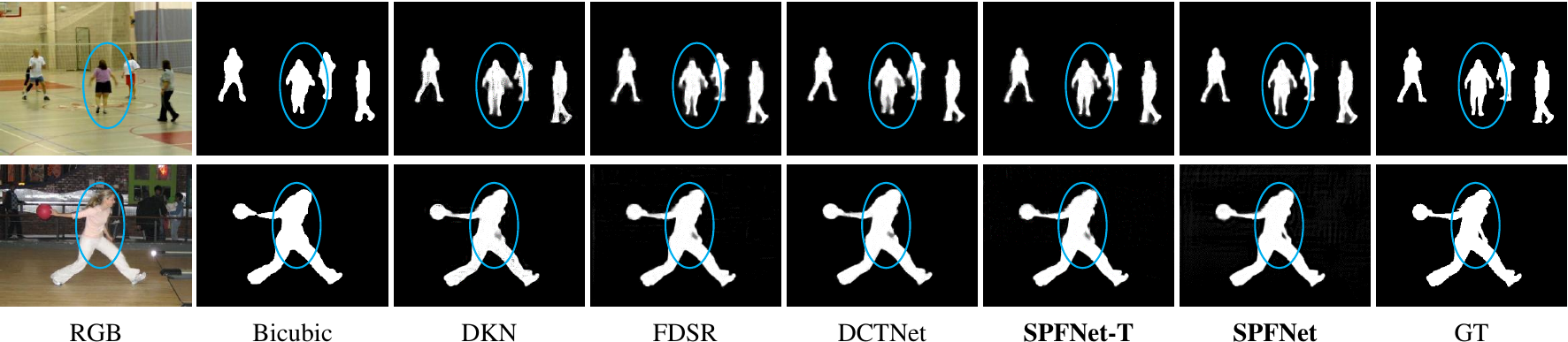}
\caption{Visual results of saliency map super-resolution on the DUT-OMRON dataset ($\times 8$).}
\label{fig:SaliencyMap}
\end{figure*}

\noindent \textbf{Model Complexity Analysis.} To fairly validate the effectiveness of our method, Fig.~\ref{fig:Params_Time_X16Lu} presents a comparative analysis between SPFNet-D and previous state-of-the-art methods under identical experimental settings, evaluating performance, parameters, memory usage, inference time, and FLOPs. It can be clearly observed that our method achieves optimal performance while maintaining competitive computational costs. For example, compared with the second-best SGNet, our method achieves a $9.30\%$ reduction in RMSE while maintaining comparable memory usage, and significantly decreases the number of parameters by $64.75\%$, inference time by $46.06\%$, and FLOPs by $68.59\%$.

\noindent \textbf{Joint DSR and Denoising.}
Tab.~\ref{tab:denoising_NYU} demonstrates that our method outperforms other approaches in joint DSR and denoising on four benchmark datasets using bicubic downsampling. Depth obtained in real-world environments is often noisy, which poses a challenge to HR depth restoration. 
Similar to the existing methods~\cite{kim2021deformable,zhong2023deep,shin2023task}, we first add Gaussian noise (mean 0 and standard deviation 0.07) to the LR depth as a new input. As evidenced by Tab.~\ref{tab:denoising_NYU} (top), compared to the second-best method, SPFNet reduces the RMSE ($\times 16$) by $1.23cm$ on the NYU-v2 dataset and by $0.68cm$ on the Middlebury dataset.  Fig.~\ref{fig:Denoising} (Noisy depth) presents the visual comparison on the $\times8$ Middlebury dataset. Notably, both our SPFNet and SPFNet-D demonstrate strong noise robustness, enabling them to recover highly precise and sharp depth from noisy environments. For instance, the edges of the toy and the cord restored by our method are clearer compared to other approaches. 

Following the previous method~\cite{shin2023task}, we further introduce Gaussian noise to both the LR depth and RGB images as new inputs to simulate challenging real-world environments. As shown in Tab.~\ref{tab:denoising_NYU} (bottom), our SPFNet surpasses the suboptimal approach ($\times 16$) by $4.19cm$ RMSE on the NYU-v2 dataset and by $1.19cm$ RMSE on the Lu dataset. In addition, although our SPFNet-D exhibits a performance drop compared to the original SPFNet, it still achieves results comparable to those of recent advanced methods (\textit{e.g.}, SUFT~\cite{shi2022symmetric}, DCTNet~\cite{zhao2022discrete}, and SGNet~\cite{wang2024sgnet}) without incurring the additional computational cost of large-scale models. As illustrated in Fig.~\ref{fig:Denoising} (Noisy RGB-D), the edges of the bowling pin reconstructed by our method are visibly more accurate and closer to the GT depth. In summary, these results thoroughly confirm the robustness of our method.

\begin{table*}[t]
\caption{Quantitative comparisons of depth completion on the real-world TOFDC dataset.}\label{tab:dc}
	\centering
    \Huge
	\resizebox{1\linewidth}{!}{
\begin{tabular}{c|ccccccccccc}
\toprule 
Metrics  &CSPN \cite{2018Learning} &FusionNet \cite{van2019sparse} & GuideNet \cite{tang2020learning} &NLSPN \cite{park2020non}   	&CFormer \cite{zhang2023cf} 	&RigNet \cite{yan2022rignet} &PointDC \cite{yu2023aggregating} &TPVD \cite{yan2025tri} &\textbf{SPFNet-D}&\textbf{SPFNet}\\
\midrule
RMSE $\downarrow$                    &22.4     &11.6              &14.6      &17.4     &11.3            &13.3                   &10.9     &9.2     &\underline{8.1}      &\textbf{8.0}	\\
REL $\downarrow$                     &0.042    &0.024             &0.030     &0.029    &0.029           &0.025                  &0.021    &\underline{0.014}   &\textbf{0.013}    &\underline{0.014}	\\
$\delta _{1.25} $ $\uparrow$   	     &94.5     &98.3              &97.6      &96.4     &\underline{99.1}            &97.6                   &98.5     &\underline{99.1}    &\textbf{99.2}     &\textbf{99.2}	\\
$\delta _{1.25^{2} } $ $\uparrow$    &95.3     &99.4              &98.9      &97.9     &\underline{99.6}            &99.1                   &99.2     &\underline{99.6}    &\underline{99.6}     &\textbf{99.7}	\\
$\delta _{1.25^{3} } $  $\uparrow$   &96.5     &99.7  &99.5      &98.9     &\textbf{99.9}   &99.7      &99.6     &\textbf{99.9}       &\underline{99.8}     &\textbf{99.9}	\\
\bottomrule
\end{tabular}}
\end{table*}

\begin{figure*}[t]
\centering
\includegraphics[width=1\linewidth]{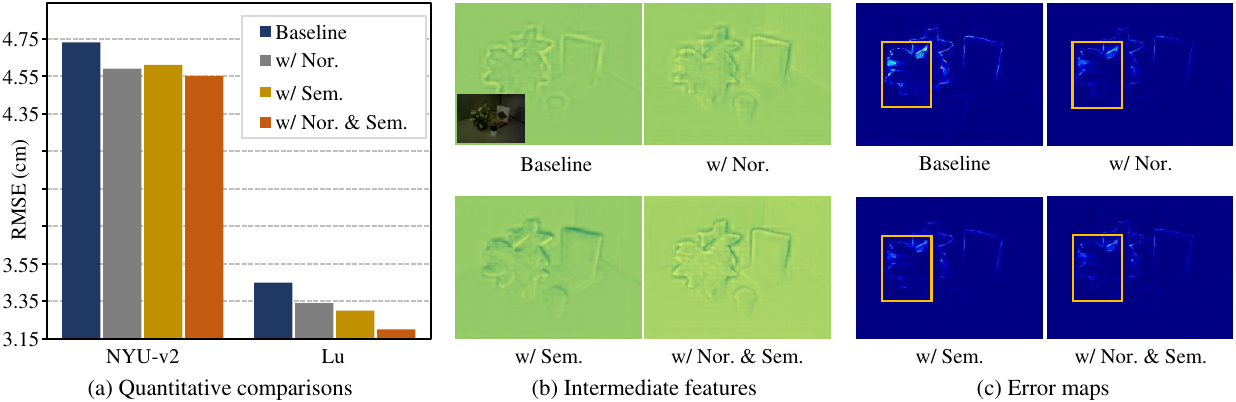}
\caption{Ablation study of surface normal and semantic on $\times 16$ DSR. The features and error maps are derived from Lu.}
\label{fig:Ablation_priors}
\end{figure*}

\begin{table*}[t]
\caption{Complexity comparisons on NYU-v2 and Lu ($\times 16$), where inference time is tested on NYU-v2.}\label{tab:ablation_effect} 
\large
\centering
\renewcommand\arraystretch{1.05}
\resizebox{0.99\linewidth}{!}{
\begin{tabular}{l|cccc|cccc}
\toprule 
\multirow{2}{*}{Methods} & \multicolumn{4}{c|}{w/o Nor. \& Sem.} & \multicolumn{4}{c}{w/ Nor. \& Sem.} \\
 \cmidrule{2-9}
&NYU-v2 &Lu &Params (M) &Time (ms) &NYU-v2 &Lu &Params (M) &Time (ms)\\
 \midrule
FDSR~\cite{he2021towards}     &5.86  &5.00  &0.60  &\textbf{14.13}    &5.76 $(-0.10)$ &4.59 $(-0.41)$ &0.67 $(+0.07)$ &\textbf{14.57} $(+0.44)$ \\
SUFT~\cite{shi2022symmetric}  &4.86  &3.92  &94.36 &15.65             &4.81 $(-0.05)$ &4.07 $(+0.15)$ &99.25 $(+4.89)$   &28.58 $(+12.93)$ \\
SGNet~\cite{wang2024sgnet}    &4.77  &3.55  &85.94 &88.81             &4.70 $(-0.07)$ &3.28 $(-0.27)$ &86.89 $(+0.95)$   &102.35 $(+13.54)$  \\
\textbf{SPFNet-T}             &5.80  &4.44   &\textbf{0.57}  &26.47   &5.71 $(-0.09)$ &4.31 $(-0.13)$ &\textbf{0.65} $(+0.08)$ &29.57 $(+3.10)$\\
\textbf{SPFNet}               &\textbf{4.73}  &\textbf{3.45}  &26.99  &34.02    &\textbf{4.55} $(-0.18)$ &\textbf{3.20} $(-0.25)$ &31.10 $(+4.11)$   &51.59 $(+17.57)$    \\
\bottomrule
\end{tabular}}
\end{table*}

\subsection{Generalization on Other Restoration Tasks}
To thoroughly validate the generalization capability of our method, we further conduct experiments on other guided image restoration tasks, including pan-sharpening, saliency map super-resolution, and depth completion, while keeping the overall architecture of our method unchanged.

\noindent \textbf{Pan-Sharpening.}
Tab.~\ref{tab:pan} reports the superiority of SPFNet over advanced Pan-Sharpening methods on WorldView III and GaoFen2 datasets, including GFPCA \cite{liao2015two}, PanNet \cite{yang2017pannet}, MSDCNN \cite{yuan2018multiscale}, SRPPNN \cite{cai2020super}, GPPNN \cite{xu2021deep}, MutInf \cite{zhou2022mutual}, and PanFlow \cite{yang2023panflownet}. Since the ground truth itself is not available, we follow previous methods \cite{yuan2018multiscale,yang2023panflownet,zhou2022mutual} to generate synthetic datasets by using the Wald protocol tool~\cite{wald1997fusion}. 
In the training and test sets, PAN images and LR multi-spectral images are cropped to sizes of $128 \times 128$ and $32 \times 32$, respectively. Additionally, PSNR, SSIM, SAM \cite{yuhas1992discrimination}, and ERGAS \cite{alparone2007comparison} are employed as image quality assessment metrics to evaluate our experimental results. Compared to the suboptimal method, we can clearly see that our SPFNet increases the PSNR by $0.1879dB$ on the WorldView III dataset and $0.1776dB$ on the WGaoFen2 dataset.

\begin{table*}[t]
\caption{Ablation study of SPFNet with different normal and semantic models on $\times16$ DSR.}\label{tab:Ablation_diffModels} 
\small
\centering
\renewcommand\arraystretch{1.2}
\resizebox{0.8\linewidth}{!}{
\begin{tabular}{l|cc|cc|cc}
\toprule 
\multirow{2}{*}{Methods}   & \multicolumn{2}{c|}{Semantic models} & \multicolumn{2}{c|}{Normal models}   & \multicolumn{2}{c}{Datasets}\\
 \cmidrule{2-7}
&MobileSAM~\cite{zhang2023faster}           &SAM~\cite{kirillov2023segment}           &Metric3Dv2~\cite{hu2024metric3d}     &Omnidata~\cite{eftekhar2021omnidata}         &NYU-v2           &Lu                \\
\midrule
(a)      &\checkmark              &             &\checkmark             &            &4.56            &3.30               \\
(b)      &\checkmark              &             &             &\checkmark            &4.61            &3.33               \\
(c)      &              &\checkmark             &\checkmark             &            &4.56            &3.25               \\
(d)      &              &\checkmark             &             &\checkmark            &\textbf{4.55}            &\textbf{3.20}               \\
\bottomrule
\end{tabular}}
\end{table*}

\begin{figure*}[t]
\centering
\includegraphics[width=0.98\linewidth]{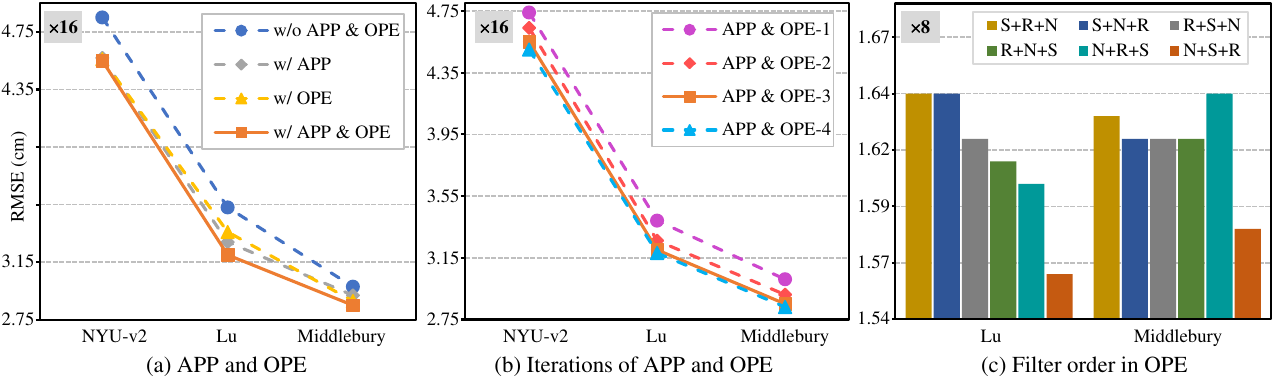}
\caption{Ablation study of APP and OPE. S+R+N refers to the filter in the order of semantic (S), RGB (R), and normal (N).} 
\label{fig:Ablation_APP_OPE}
\end{figure*}

\noindent \textbf{Saliency Map Super-resolution.}
Tab.~\ref{tab:saliencysr} performs $\times4$, $\times8$, and $\times16$ experiments on the DUT-OMRON~\cite{yang2013saliency} dataset, which comprises $5,168$ pairs of RGB and saliency maps, where the LR saliency maps are generated by downsampling the GT saliency maps using bicubic interpolation. Following~\cite{kim2021deformable,zhong2023deep}, the pre-trained model on the NYU-v2 dataset is directly tested on DUT-OMRON without any fine-tuning. Besides, we employ the F-measure as an evaluation metric to maintain consistency with previous methods. Tab.~\ref{tab:saliencysr} shows that our method achieves excellent performance. For example, our SPFNet surpasses the suboptimal approach by 0.0016 in the $\times8$ result. Furthermore, Fig.~\ref{fig:SaliencyMap} presents the visual results, showing that our method can predict high-quality HR saliency maps with clearer edges than others.


\noindent \textbf{Depth Completion.} Tab.~\ref{tab:dc} compares our method with previous state-of-the-art depth completion (DC) approaches on the real-world TOFDC~\cite{yan2024tri} dataset, including CSPN \cite{2018Learning}, FusionNet \cite{van2019sparse}, GuideNet \cite{tang2020learning}, NLSPN \cite{park2020non}, CFormer \cite{zhang2023cf}, RigNet \cite{yan2022rignet}, PointDC \cite{yu2023aggregating}, and TPVD \cite{yan2025tri}. Consistent with these DC methods, we select RMSE, REL, and $\delta _{1.25^x} (x=1,2,3)$ as evaluation metrics. It can be observed that both our SPFNet and SPFNet-D achieve satisfactory performance across all evaluation metrics, surpassing methods specifically designed for the DC task. For example, compared with the suboptimal TPVD \cite{yan2025tri} and PointDC \cite{yu2023aggregating}, our SPFNet reduces RMSE by $1.2cm$ and $2.9cm$, respectively.

\subsection{Ablation Studies}
In this section, all ablation studies are conducted based on SPFNet on the synthetic dataset with bicubic downsampling.

\noindent \textbf{Surface Normal and Semantic priors.}
Fig.~\ref{fig:Ablation_priors} shows the ablation results on surface normal and semantic priors. For the baseline, similar to~\cite{he2021towards,zhao2022discrete,wang2024sgnet}, only RGB is utilized as guidance. From Fig.~\ref{fig:Ablation_priors}(a), we find that both surface normal and semantic priors contribute to a decrease in RMSE. When both are employed together, SPFNet achieves the best performance. For example, compared to the baseline, surface normal and semantic priors reduce the RMSE by $0.11cm$ and $0.15cm$ on the Lu dataset, respectively. Finally, SPFNet outperforms the baseline by $0.25cm$ on the Lu dataset.

Fig.~\ref{fig:Ablation_priors}(b) and (c) illustrate the visual results of intermediate depth features and error maps. We observe that both surface normal and semantic priors are capable of producing more distinctive edges and fewer errors than the baseline. Furthermore, when combining them, our SPFNet generates a much clearer structure and the much lesser errors. These results suggest that, by leveraging scene priors, our SPFNet can effectively enhance depth edges and minimize errors.

Tab.~\ref{tab:ablation_effect} provides the comparison of incorporating surface normal and semantic priors into previous advanced methods, further validating the effectiveness of our SPFNet. Specifically, for previous approaches, we add additional branches for both surface normal and semantic priors. These branches share the same architecture as the RGB branch present in their original networks. For one thing, as demonstrated by the second column (w/o Nor. \& Sem.) in Tab.~\ref{tab:ablation_effect}, our SPFNet achieves the best performance and competitive complexity even without relying on surface normal and semantic priors. For example, SPFNet is $0.1cm$ lower compared to the suboptimal methods on the Lu dataset. For another thing, when compared to the second column, the third column (w/ Nor. \& Sem.) in Tab.~\ref{tab:ablation_effect} indicates that nearly all DSR approaches benefit from integrating surface normal and semantic priors. Notably, our method still outperforms others. For instance, compared to the second-best method (SGNet), our SPFNet not only exhibits lower trainable parameters and inference time, but it also decreases the RMSE by $0.15cm$ on the NYU-2 dataset. In short, whether surface normal and semantic priors are employed or not, our method consistently delivers superior performance.

\begin{table*}[t]
\caption{Ablation study of MGF on NYU-v2 and Lu datasets ($\times8$). D2P: depth-to-prior filtering, P2D: prior-to-depth filtering. D2P$\rightarrow$P2D means that the D2P step precedes the P2D step.}\label{tab:Ablation} 
\centering
\LARGE
\renewcommand\arraystretch{1.2}
\resizebox{1\linewidth}{!}{
\begin{tabular}{l|cc|cccc|c|c|cc}
\toprule 
Methods   &D2P          &P2D          &D2P$\rightarrow$D2P  &P2D$\rightarrow$P2D    &D2P$\rightarrow$P2D      &P2D$\rightarrow$D2P     &Similarity  &Params (M)         &NYU-v2           &Lu                \\
\midrule
(a)      &              &             &             &            &           &            &            &29.76 $(\pm0.00)$  &2.59 $(\pm0.00)$ &1.75 $(\pm0.00)$              \\
(b)      &\checkmark    &             &             &            &           &            &            &30.22 $(+0.46)$    &2.51 $(-0.08)$   &1.70 $(-0.05)$              \\
(c)      &              &\checkmark   &             &            &           &            &            &30.22 $(+0.46)$    &2.48 $(-0.11)$   &1.64 $(-0.11)$ 	         \\ \midrule
(d)      &\checkmark    &             &\checkmark   &            &           &     &                   &30.68 $(+0.92)$    &2.41 $(-0.18)$   &1.60 $(-0.15)$                              \\
(e)      &              &\checkmark   &             &\checkmark  &           &&                        &30.68 $(+0.92)$    &2.39 $(-0.20)$   &1.64 $(-0.11)$                              \\              
(f)      &\checkmark    &\checkmark   &             &            &\checkmark &   &                     &30.68 $(+0.92)$    &2.46 $(-0.13)$   &1.61 $(-0.14)$                              \\
(g)      &\checkmark    &\checkmark   &             &            &           &\checkmark &             &30.68 $(+0.92)$    &2.40 $(-0.19)$   &1.59 $(-0.16)$                             \\
(h)      &\checkmark    &\checkmark   &             &            &           &\checkmark  &\checkmark  &30.68 $(+0.92)$    &\textbf{2.36} $(-0.23)$   &\textbf{1.56} $(-0.19)$              \\
\bottomrule
\end{tabular}}
\end{table*}

\noindent \textbf{Different Large-Scale Models.} Tab.~\ref{tab:Ablation_diffModels} reveals the ablation study on different combinations of large-scale models. The results demonstrate the strong compatibility of our SPFNet with various foundation models, as all four configurations achieve satisfactory performance. Specifically, with the semantic model fixed, using Omnidata as the normal model yields a slight performance improvement over Metric3Dv2. Conversely, when the normal model is fixed, SAM exhibits certain advantages over MobileSAM. Finally, we adopt configuration (d) as the default setup for SPFNet, which surpasses configurations (a)-(c) on the Lu dataset by RMSE margins of $0.10cm$, $0.13cm$, and $0.05cm$, respectively.

\begin{figure}[t]
\centering
\includegraphics[width=1\linewidth]{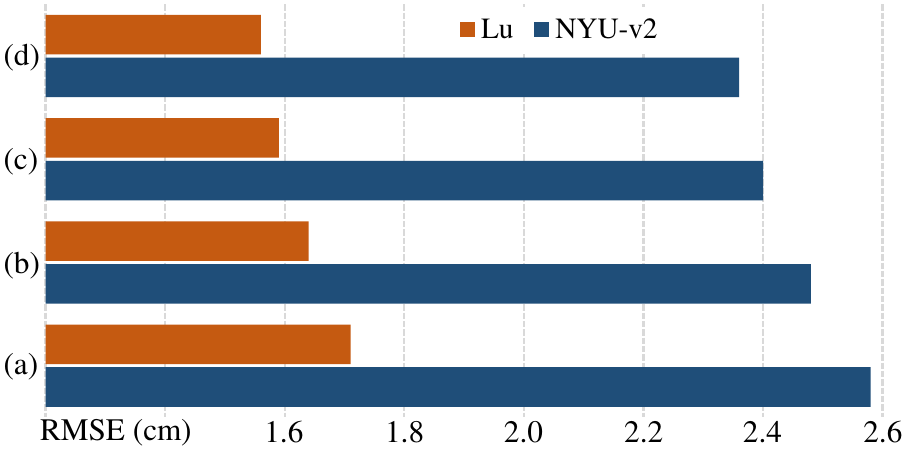}
\caption{Ablation study on the effectiveness of similarity computation on the NYU-v2 and Lu datasets ($\times8$).}
\label{fig:similarity_effectiveness}
\end{figure}

\noindent \textbf{APP and OPE.}
Fig.~\ref{fig:Ablation_APP_OPE}(a) presents the ablation on APP and OPE. The baseline (blue dotted line) removes all APP and OPE modules from SPFNet. It can be discovered that both our APP and OPE contribute to performance improvement. When they are utilized in conjunction, our method (solid orange line) achieves the best performance. For instance, our SPFNet surpasses the baseline by $0.3cm$ on the NYU-v2 dataset and $0.33cm$ on the Lu dataset, respectively.

In addition, Fig.~\ref{fig:Ablation_APP_OPE}(b) shows the ablation on different iterations of APP and OPE modules. It is evident that the performance incrementally improves as the number of APP and OPE increases. When APP \& OPE-4 (4 iterations of APP and OPE) is utilized, the RMSE consistently reduces on the NYU and Lu datasets, but only slightly on the Middlebury dataset. To better balance complexity and performance, we select APP \& OPE-3 (orange line) as the setting for SPFNet.


Finally,  Fig.~\ref{fig:Ablation_APP_OPE}(c) presents an ablation study investigating the ordering of single-modal prior filters within the OPE module on the Lu and Middlebury datasets. These results clearly demonstrate the superiority of the ‘N+S+R’ configuration, which achieves the lowest RMSE.

\noindent \textbf{MGF.}
Tab.~\ref{tab:Ablation} reports the ablation study of MGF on $\times 8$ DSR. For the baseline (a), we remove all of the guided image filtering blocks in SPFNet. (b) and (c) employ solely depth-to-prior filtering (D2P) or prior-to-depth filtering (P2D), respectively, both of which enhance DSR performance compared to (a). (d)-(g) explore different combination orders of D2P and P2D, and the results show that these combinations achieve lower RMSE than using D2P or P2D alone. Based on (g), (h) further leverages the similarity weights to guide the generation of filter kernels, contributing to the best performance, \textit{i.e.}, surpassing the baseline (a) by $0.23cm$ on the NYU-v2 dataset and $0.19cm$ on the Lu dataset, respectively.

\begin{figure}[t]
\centering
\includegraphics[width=1\linewidth]{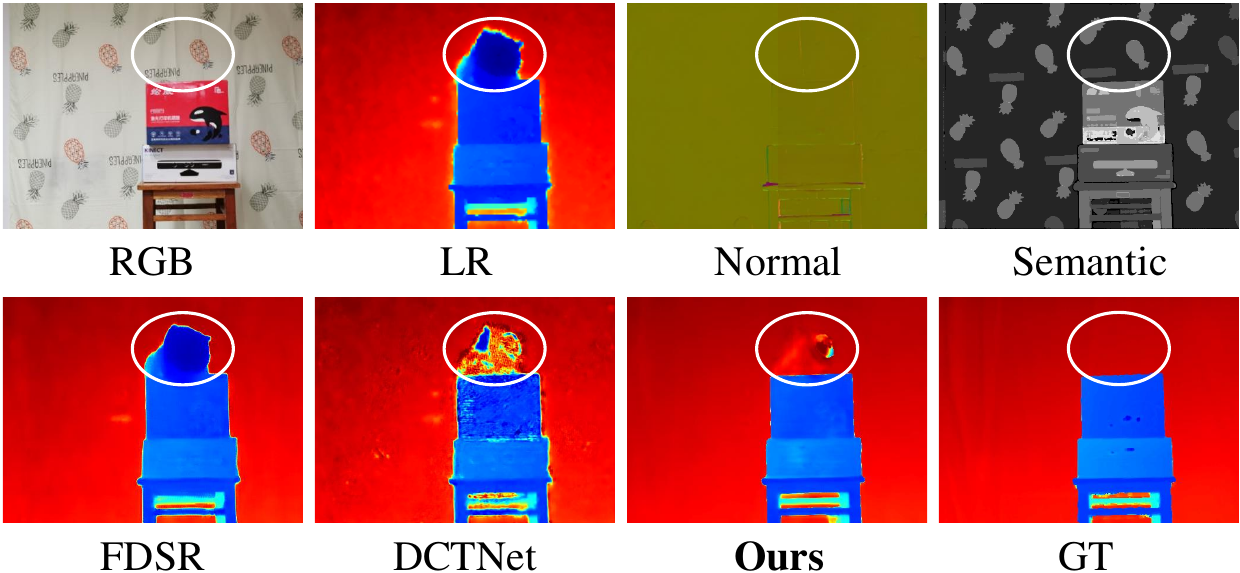}
\caption{Failure cases resulting from an inaccurate prior.} 
\label{fig:FailureCase}
\end{figure}

\noindent \textbf{Effectiveness of Similarity Computation.} Fig.~\ref{fig:similarity_effectiveness} illustrates the ablation study on multimodal similarity computation. (a) is the baseline, where the entire APP and similarity weights input in OPE are removed. (b) removes similarity computation in APP and similarity weights in OPE. (c) retains full APP but removes similarity weights in OPE. (d) uses full APP and OPE, \textit{i.e.}, our default setting. These results demonstrate that similarity computation yields satisfactory performance gains, surpassing the baseline method (a) by $0.22cm$ in RMSE on the NYU-v2 dataset.

\section{Conclusion}
In this paper, we propose SPFNet, a novel DSR solution that utilizes the surface normal and semantic priors from large-scale models to effectively weaken the texture interference and improve the edge accuracy.
Specifically, we design an all-in-one prior propagation that mitigates interference by calculating the similarity weights between multi-modal scene priors. 
Moreover, we develop the one-to-one prior embedding that continuously aggregates each single-modal prior into the depth using mutual guidance filtering, further reducing interference and enhancing the edge. 
Comprehensive experiments demonstrate that our SPFNet performs favorably against state-of-the-art approaches.

\section{Discussion}

\noindent\textbf{Limitation.} Our method utilizes high-quality prior knowledge from large models to reduce texture interference in RGB and enhance edges, thereby significantly improving the model's representation capability. However, when the input data falls outside the distribution of the large model's training samples, the priors may introduce noise. Such errors can potentially interfere with subsequent similarity computation and mutual filtering, degrading performance.

\noindent\textbf{Failure Case.} As depicted in Fig.~\ref{fig:FailureCase}, when the input RGB falls outside the distribution of the large model's training samples, the generated priors may not be sufficiently accurate (\textit{e.g.}, blurred normal structures and depth-independent semantic information). In such cases, although our method achieves higher recovery quality than other approaches, it still struggles to precisely restore regions in real scenes that exhibit severe structural distortions and artifacts. For example, the artifacts shown in the white areas in Fig.~\ref{fig:FailureCase}.

The primary cause of this failure lies in the severe degradation inherent in real-world depth maps, such as structural distortion, artifacts, and holes. When the predicted scene prior is also inaccurate, the compounded errors lead to significant deviations in the computed similarity. These deviations cause the subsequent mutual guidance filter to transfer incorrect structural priors into the depth, thereby interfering with the depth reconstruction.

A potential solution is to build a collaborative optimization framework that jointly fine-tunes normal estimation, semantic segmentation, and DSR to improve the accuracy of scene priors. Besides, the large model could serve as an intermediate supervision signal rather than directly incorporating its outputs. This strategy will mitigate the adverse effects of low-quality priors in the inference stage.


\begin{acknowledgements}
The authors would like to thank the editor and the anonymous reviewers for their critical and constructive comments and suggestions. This work was supported by National Natural Science Foundation of China under Grant Nos. U24A20330, 62361166670 and 62406135, and Natural Science Foundation of Jiangsu Province under Grant No. BK20241198.
\end{acknowledgements}

\section*{Data availability statements}
Information on access to the datasets supporting the conclusions of this article is included therein.

%
\section*{Conflict of interest}

The authors declare that they have no conflict of interest.


\bibliographystyle{spmpsci}      
\bibliography{egbib.bib}   

\end{document}